\documentclass[runningheads]{llncs}
\usepackage{graphicx}
\usepackage{comment}
\usepackage{amsmath,amssymb} 
\usepackage{color}
\usepackage{xcolor}
\usepackage{wrapfig}
\usepackage{pifont}

\usepackage{mathrsfs}
\usepackage{microtype}
\usepackage{booktabs}

\usepackage[linesnumbered,ruled,vlined]{algorithm2e}
\SetKwInput{KwInput}{Input}                
\SetKwInput{KwOutput}{Output}              

\SetCommentSty{mycommfont}

\usepackage{booktabs} 
\usepackage{caption}
\usepackage{multirow}

\usepackage[width=122mm,left=12mm,paperwidth=146mm,height=193mm,top=12mm,paperheight=217mm]{geometry}

\newcommand{\cmark}{\color{green}\ding{51}}%
\newcommand{\xmark}{\color{red}\ding{55}}%

\newcommand{\etal}{\textit{et al.}}

\newcommand{\hidden}[1]{}

\newcommand{\ignore}[1]{}
\newcommand{\todo}[1]{}
\newcommand{\methodname}{GSDN\xspace}

\DeclareMathOperator\supp{supp}

\begin{document}
\pagestyle{headings}

\title{Generative Sparse Detection Networks \\ for 3D Single-shot Object Detection}


\titlerunning{Generative Sparse Detection Networks for 3D Single-shot Object Detection}
%
\author{JunYoung Gwak\inst{1} \and
Christopher Choy\inst{2} \and
Silvio Savarese\inst{1}}
\authorrunning{J. Gwak et al.}
%
\institute{Stanford University
\email{\{jgwak,ssilvio\}@stanford.edu} \and
NVIDIA
\email{cchoy@nvidia.com}}
\maketitle

\begin{abstract}
3D object detection has been widely studied due to its potential applicability to many promising areas such as robotics and augmented reality. Yet, the sparse nature of the 3D data poses unique challenges to this task. Most notably, the observable surface of the 3D point clouds is disjoint from the center of the instance to ground the bounding box prediction on. To this end, we propose Generative Sparse Detection Network (GSDN), a fully-convolutional single-shot sparse detection network that efficiently generates the support for object proposals. The key component of our model is a generative sparse tensor decoder, which uses a series of transposed convolutions and pruning layers to expand the support of sparse tensors while discarding unlikely object centers to maintain minimal runtime and memory footprint. GSDN can process unprecedentedly large-scale inputs with a single fully-convolutional feed-forward pass, thus does not require the heuristic post-processing stage that stitches results from sliding windows as other previous methods have. We validate our approach on three 3D indoor datasets including the large-scale 3D indoor reconstruction dataset~\cite{dai2017scannet} where our method outperforms the state-of-the-art methods by a relative improvement of 7.14\% while being 3.78 times faster than the best prior work.
\keywords{Single shot detection, 3D object detection, generative sparse network, point cloud}
\end{abstract}

\section{Introduction}

3D reconstructions have become more commonplace as a complete reconstruction pipeline become built into consumer devices, such as mobile phones or head-mounted displays, for applications in robotics and augmented reality. 
Among these applications, perceptions on 3D reconstructions is the first step allowing users to interact with a virtual world in 3D. For example, indoor navigation applications can aid a user to localize objects, and mixed reality applications need to track objects to give users information relevant to the current status of their surroundings.
Many of these virtual-reality and mixed-reality applications require identifying and detecting 3D objects in real-time.






However, unlike 2D images where the input is in a densely packed array, 3D data is scanned or reconstructed as a set of points or a triangular mesh. These data occupy a small portion of the 3D space and pose unique challenges for 3D object detection. First, the space of interest is three dimensional which requires cubic complexity to save or process data. Second, the data of interest is very sparse, and all information is sampled from the surface of objects. 



Many previous 3D object detectors proposed various methods to process cubically growing sparse 3D data, and can be categorized into one of two branches: 3D object detection by converting sparse 3D data into a dense representation~\cite{maturana_iros_2015,DeepSlidingShapes,armeni_cvpr16,li2016vehicle,hou20193d} or by directly feeding a set of points into multi-layer perceptrons~\cite{qi2019deep,yang2019learning}. First, dense 3D representation for indoor object detection~\cite{DeepSlidingShapes,armeni_cvpr16,hou20193d} uses volumetric features which have memory and computational complexity of $O(N^3)$ where $N$ is the resolution of the space. This representation requires large memory, which prevents the utilization of deep networks and requires cropping the scenes and stitching the results to process large or high-resolution scenes. Second, multi-layer perceptrons that process a scene as a set of points limit the number of points a network can process. Thus, as the size of the point cloud increases, the method suffers from either low-resolution input which makes it difficult to scale the method up for larger scenes (see Section~\ref{sec:analysis}) or apply sliding-window style cropping and stitching which prevents the network to see a larger context~\cite{yang2019learning}.

We instead propose to resolve the cubic complexity with our hierarchical sparse tensor encoder, adopting a sparse tensor network to efficiently process a large scene fully-convolutionally. As we use a sparse representation, our network is fast and memory-efficient compared with a single-shot method that uses dense tensors~\cite{hou20193d}. It allows our network to adopt extremely deep architectures while requiring a fraction of the memory and computation.
Also, compared with multi-layer perceptrons, our method scales to large scenes without sacrificing point density or the receptive field size of a network by cropping a scene into smaller windows~\cite{qi2019deep,yang2019learning}. 

\begin{figure}[t]
    \centering
    \small
    \includegraphics[width=.99\linewidth]{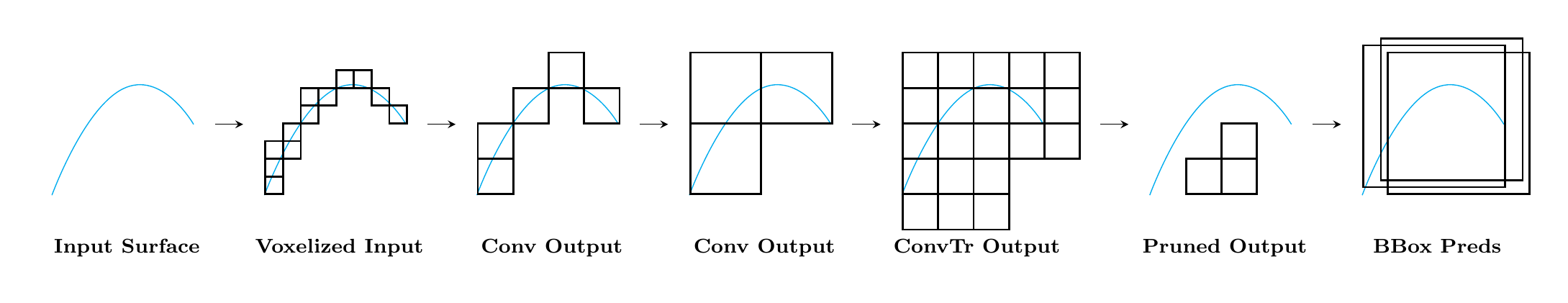}
    \caption{The top-down view of the cross-section of our simplified 3D sparse anchor generation pipeline: a 3D scanner samples the surface of an object which we convert to a sparse tensor. Then, an encoder extracts hierarchical sparse tensor features with a series of convolutions. During the decoder stage, we apply a transposed convolution to upsample and expand the support of the sparse tensor. Finally, we prune out unnecessary supports that do not contain anchors and make bounding box anchor predictions.}
    \label{fig:anchor_generation}
    \vspace{-1em}
\end{figure}

Another key challenge of a 3D object detector is that the support of the input 3D scans and the support of the object bounding box anchors are disjoint. In other words, we have samples of 3D points on the surface of the objects, but not on the center of the object where a bounding box anchor is located. This is due to the fact that many objects are convex and we cannot directly observe the object center. 
For this, we propose a generative sparse tensor decoder that repeatedly upsamples the support of input to expand and cover the support of anchors while discarding unlikely object centers to maintain minimal runtime and memory footprint (Fig.~\ref{fig:anchor_generation}).


\begin{figure}[t]
    \centering
    \includegraphics[width=\textwidth]{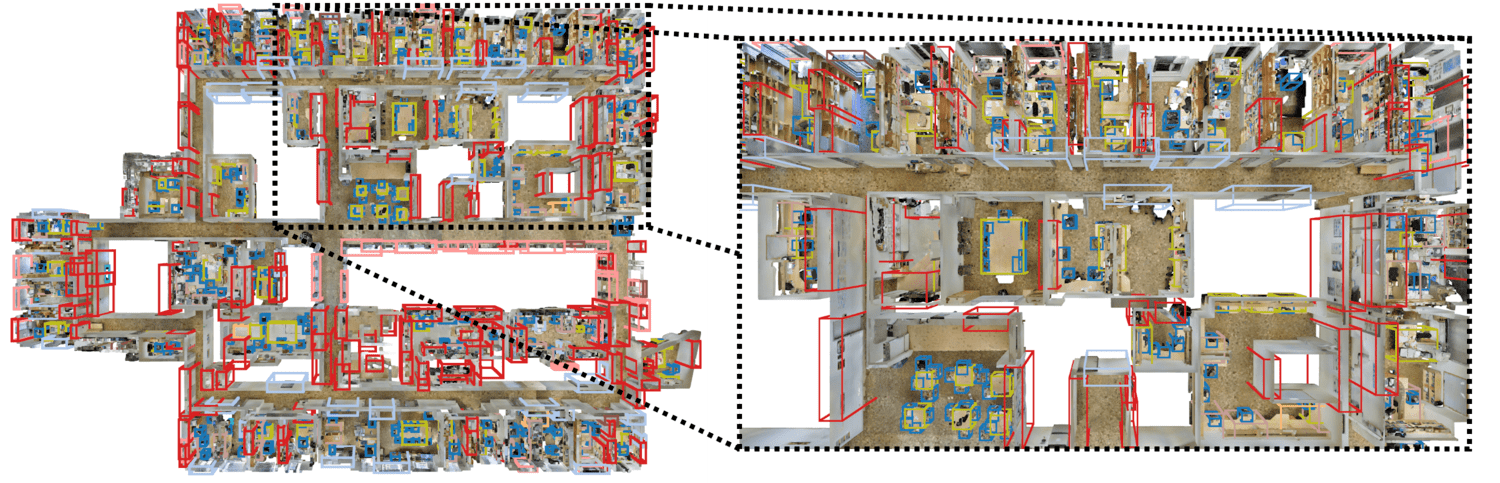}
    \caption{Detection results on the entire S3DIS building 5: Our proposed method can process 78M points, 13984m${}^3$, 53 room building as a whole in a \textit{single fully-convolutional feed-forward pass}, only using 5G of GPU memory. Left: bird-eye-view of the entire building 5, Right: partial view of the same building.}
    \label{fig:stanford_full}
    \vspace{-1em}
\end{figure}


To sum, we propose Generative Sparse Detector Network (\methodname), a deep \textit{fully-convolutional} single-shot 3D object detection algorithm with a sparse tensor network. Our single-shot 3D object detection network consists of two components: an hierarchical sparse tensor encoder which efficiently extracts deep hierarchical features, and a generative sparse tensor decoder which expands the support of the sparse input to ground object proposals on. Experimentally, \methodname outperforms the state-of-the-art methods on two large-scale indoor datasets while being faster than the best prior work. We also analyze the speed and memory footprint of the model and demonstrate the extreme scalability of our method on orders of magnitudes larger 3D scenes (Fig. ~\ref{fig:stanford_full}).




\section{Related Work}

In this section, we review a few branches that are related to our work: 3D indoor object detection, 3D generative networks, and sparse tensor networks.


\noindent\textbf{3D Indoor Object Detection.}
In a 3D indoor setting or 3D indoor datasets~\cite{dai2017scannet,armeni_cvpr16}, the distribution of object placement creates unique challenges: objects such as lamps and ceiling lights can be placed on a wall or a ceiling, or objects can be placed on top of another object such as a desk or a bed. However, such challenging setup does not exist in outdoor datasets and most 3D outdoor object detectors simply project the 3D problem into a 2D ground plane~\cite{maturana_iros_2015,li2016vehicle,Zhou_2018_CVPR}.

Thus, in this section, we cover 3D indoor object detection specifically. The indoor 3D object detection using neural networks can be classified into one of the following categories: sliding-window with classification, clustering-based methods, bounding-box proposal, or combinations of the above methods. 
First, the sliding window with classification extracts a 3D patch for object classification which is used as a simple object detector~\cite{DeepSlidingShapes,armeni_cvpr16}\hidden{Deep Sliding Shapes}.

Second, clustering-based methods learn features or vectors in a metric space where clustering results in instance segmentation. Lahoud~\etal~\cite{lahoud20193d}\hidden{MTML} uses metric learning to train the feature space. Liu~\etal~\cite{liu2019masc}\hidden{MASC}, Yi~\etal~\cite{yi2019gspn}\hidden{GSPN}, Wang~\etal~\cite{wang2018sgpn}\hidden{SGPN}, and Qi~\etal~\cite{qi2019deep}\hidden{Hough voting~\cite{qi2019deep}} predict object centers per 3D point and cluster the center votes.

Third, the bounding box proposal methods adopt 2D rectangular bounding box proposal methods to 3D. Wang~\etal~\cite{wang2015voting} proposed Vote3D, which predicts 3D bounding boxes on a sparse grid for object detection. Yang~\etal~\cite{yang2019learning} directly predicts bounding boxes from MLP of global point cloud features. Hou~\etal~\cite{hou20193d} makes a straight-forward 3D extension of region proposal networks on dense voxels. \methodname is a bounding box proposal method with a crucial difference in maintaining the sparsity of the input point cloud and target anchor space, enabling much faster inference on many orders of magnitude larger scene with better performance than state-of-the-art methods.

\noindent\textbf{3D Generative Networks.} Generating 3D shapes from a neural network can be classified into two broad categories: continuous 3D point representations~\cite{Meschedar2019,Park_2019_CVPR,yuan2018pcn,topnet2019} and discrete grid representations~\cite{choy20163d,ogn2017,minkowskinet,dai2018scancomplete,dai2019sg}. Specifically, within the discrete representations, some use sparse representations for 3D reconstruction which allow a high-resolution voxel or signed-distance-function (SDF) reconstruction~\cite{ogn2017,minkowskinet,dai2018scancomplete,dai2019sg}. 

Unlike previous works that focus on the shapes of objects, we use the generative process to predict the bounding box anchors. Also, compared with some sparse generative processes that subdivide voxels~\cite{ogn2017,dai2019sg}, our method extends the support with transposed convolutions to cover bounding box anchors which are located behind 3D surface observations.


\noindent\textbf{Sparse Tensor Networks.}
A conventional neural network processes a dense tensor such as temporal data, images, or videos using a series of linear operations and non-linear operations. Most of the linear operations also use dense tensors for parametrization. Recently, using a sparse parametrization to compress a neural network~\cite{han2015deep,parashar2017scnn,narang2017exploring} has been widely studied for mobile and embedded systems. However, using a sparse tensor as an input has only gained popularity after its success on 3D data processing~\cite{sparseconvnet,SubmanifoldSparseConvNet,minkowskinet,FCGF2019}. Note that these networks are different from the compressed models using parameter pruning whose weights are sparse matrices but all feature maps are dense tensors; whereas the sparse tensor networks take \textit{spatially} sparse tensors as inputs and generate \textit{spatially} sparse feature maps. We adopt these spatially sparse networks, or sparse tensor networks to scale detection networks to an unprecedented depth and to handle extremely large scenes.

\section{Preliminaries}

In this section, we briefly go over the basic 3D representation, a sparse tensor, and introduce basic operations that are critical for the generative sparse tensor network. Throughout the paper, we will use lowercase letters for variable scalars, $t$; uppercase letters for constants, $N$; lowercase bold letters for vectors, $\mathbf{v}$; uppercase bold letters for matrices, $\mathbf{R}$; Euler scripts for tensors, $\mathscr{T}$; and calligraphic symbols for sets, $\mathcal{C}$.


\subsection{Sparse Tensor}

A tensor is a multi-dimensional array that can represent high-dimensional data. A sparse tensor of order-$D$, $\mathscr{T} \in \mathbb{R}^{N_1 \times N_2 \times ... \times N_D}$, is a $D$-dimensional array where majority of its elements are 0. Adopting the conventional sparse matrix representation, a sparse matrix can be represented as a set of non-zero coordinates $\mathcal{C} = \supp(\mathscr{T})$ where $\supp$ is the support operator, and corresponding features $\mathcal{F}$.
\begin{equation}
   \mathscr{T}[x^1_i,  x^2_i,  \cdots, x^D_i] = \begin{cases}
      \mathbf{f}_i \;\; & \text{if} \; (x^1_i,  x^2_i, \cdots, x^D_i) \in \mathcal{C} \\
      0   \;\; & \text{otherwise}
   \end{cases}
\end{equation}
where $x_d^i$ denotes $d$-th axis coordinate of the $i$-th non-zero element and $\mathbf{f}_i$ is the feature associated to the $i$-th non-zero element. These non-zero elements contain information that are equivalent to a sparse tensor $\mathcal{T} \Leftrightarrow (\mathcal{C}, \mathcal{F})$. These sets can also be converted to matrices $\mathbf{C}, \mathbf{F}$ in a COO representation.

\subsection{Sparse Tensor for 3D Data Representation}

The 3D data of interest in this work uses point clouds or meshes to represent 3D surfaces. We can represent a mesh or a point cloud as a sparse tensor by discretizing the coordinates of vertices or points. This process requires defining the discretization step size (voxel size) which is a hyperparameter that affects the performance of a neural network~\cite{FCGF2019,minkowskinet}. 

\section{Generative Sparse Detection Networks}

\begin{figure}[t]
\centering
\small
\includegraphics[width=.99\linewidth]{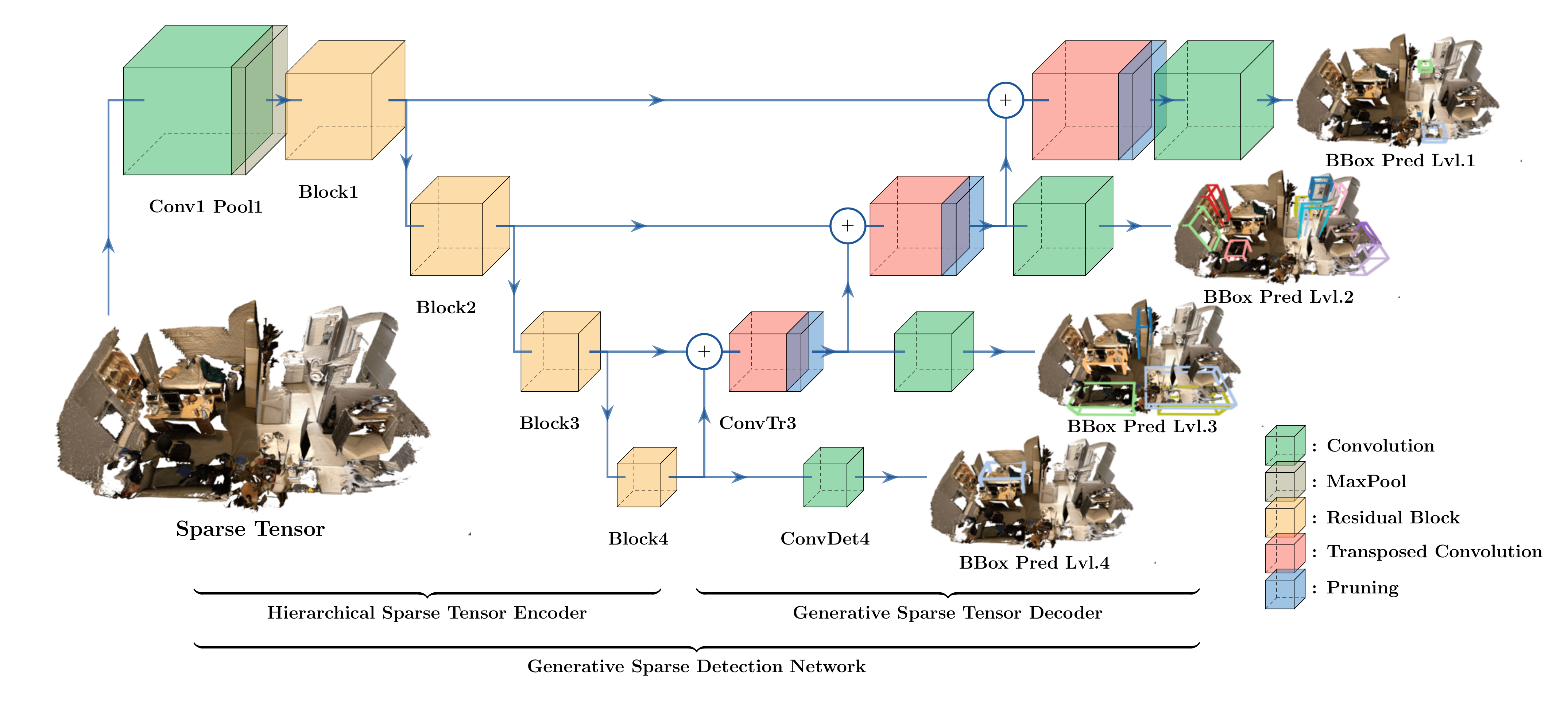}
\caption{Network overview: generative sparse detection networks process a sparse tensor input first with a series of strided convolutions followed by a few residual network blocks to generates hierarchical sparse tensor feature maps (Sec.~\ref{sec:sparse_pyramid}). The second stage upsamples the sparse tensor feature maps using transposed convolution and pruning (Sec.~\ref{sec:generative_network}). Note that all feature maps are sparse tensors and all layers process sparse tensors fully-convolutionally.}
\label{fig:network}
\vspace{-1em}
\end{figure}

In this section, we propose the generative sparse detection networks for 3D object detection. 
Unlike the 2D object detection networks~\cite{lin2017feature,ren2015faster}, we use a sparse tensor as the 3D representation throughout the network including the intermediate features. Thus, all layers such as convolution and batch normalization are well defined for sparse tensors~\cite{sparseconvnet,minkowskinet}. Throughout the paper, we will implicitly refer to all tensors as sparse tensors and layers as sparse tensor counterparts.

The network consists mainly of two parts: a hierarchical sparse tensor encoder and a generative sparse tensor decoder. The first part of the network generates sparse tensor feature maps that can sufficiently capture geometry and identity of objects and the second part proposes new supports based on the feature maps.

\subsection{Hierarchical Sparse Tensor Encoder}
\label{sec:sparse_pyramid}

We use residual networks~\cite{he2016deep}, specifically high-dimensional variants proposed in Choy~\etal~\cite{minkowskinet}, as the backbone of our model. Note that the backbone network can be replaced with more modern and recent variants. The network consists of residual blocks and strided convolutions that reduce the resolution of the space and increase the receptive field size exponentially. First, the network takes a high-resolution sparse tensor as an input $\mathscr{T}_0$ and generate hierarchical feature maps $\mathscr{T}_l$ with a series of downsampling and residual blocks $f_l(\cdot; \mathbf{W}_l)$ for $l \in [1, ..., L]$. The encoder can be represented succinctly as
$$
\mathscr{T}_l \leftarrow f_l(\mathscr{T}_{l - 1}; \mathbf{W}_l) \;\text{for}\;l\in [1,...,L]
$$
We cache all of the hierarchical sparse tensor feature maps $\mathscr{T}_l$ for $l \in [1, ..., L]$ which will be fed into the generative sparse tensor decoder.



\subsection{Generative Sparse Tensor Decoder}
\label{sec:generative_network}



The second half of the network expands the support of the hierarchical sparse tensors feature maps $\mathscr{T}_l$ to cover the support for bounding box anchors. We approximate this process with transposed convolutions (also known as upconvolution, deconvolution). Given an input sparse tensor $\mathscr{T}$, we create an output sparse tensor $\mathscr{T}'$ that $\supp(\mathscr{T}) \subset \supp(\mathscr{T}')$. However, not all voxels generated from this process contain object bounding box anchors and can be removed to limit the memory and computation cost. This process is the \textit{sparsity pruning} and we repeatedly apply a transposed convolution followed by sparsity pruning to increase the resolution of the space while limiting the memory and computation cost of a sparse tensor. During this process, we make skip connections between the hierarchical sparse tensor feature maps and the upsampled sparse tensors to recover the fine details of the input.

\subsubsection{Transposed Convolution and Sparsity Pruning}
\label{sec:pruning}

\begin{wrapfigure}{l}{0.45\textwidth}
\centering
\vspace{-0.5cm}
\includegraphics[width=.99\linewidth]{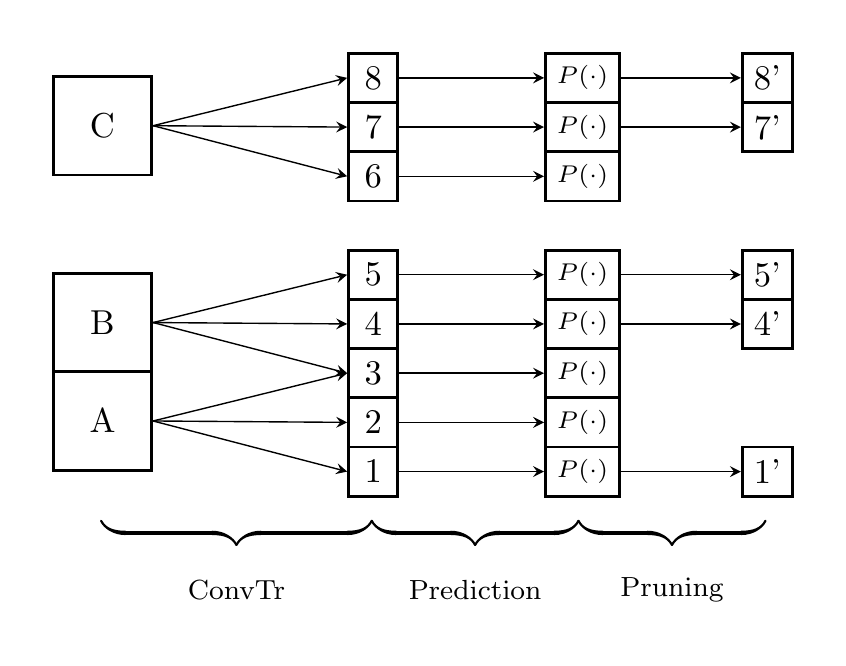}
\vspace{-0.8cm}
\caption{Expansion and pruning: transposed convolution upsamples a low-resolution sparse tensor into a high-resolution sparse tensor. Then, we prune out some of the upsampled coordinates with sparsity predictions $P_{\text{s}}(\cdot)$.}
\label{fig:pruning}
\vspace{-1cm}
\end{wrapfigure}

\todo{Emphasize dynamic coordinate generation. This is a novel contribution different from submanifold conv}
We use transposed convolutions with the kernel size greater than 2 to not just upsample, but expand the support of a sparse tensor. This process affects the sparsity pattern of a sparse tensor and the support of the output sparse tensor is the stencil or outer-product of the convolution kernel shape on the input sparsity pattern $\supp(\mathscr{T}') = \mathcal{C} \otimes [-K,...,K]^3$. Mathematically, a transposed convolution on a 3D sparse tensor $\mathscr{T}$ with $\supp(\mathscr{T}) = \mathcal{C}$ can be defined as follows:
\begin{equation}
    \mathscr{T}'[x, y, z] = \sum_{i, j, k \in \mathcal{N}(x,y,z)} \mathbf{W}[x - i, y - j, z - k] \mathscr{T}[i, j, k] \;\text{for}\;(x, y, z) \in \mathcal{C}'
\end{equation}
where $\mathcal{C}' = \mathcal{C} \otimes [-K,...,K]^3$, $\mathcal{N}(x,y,z) = \{(i,j,k)| |x - i|\le K, |y - j| \le K, |z - k| < K, (i,j,k) \in \mathcal{C}\}$, $W$ is the 3D convolution kernel weights and $2K + 1$ is the convolution kernel size. This results in denser sparsity pattern on the output tensor $\mathscr{T}'$ with $\supp(\mathscr{T}') = \mathcal{C} \otimes [-K,...,K]^3$. Note that unlike the subdivision, the transposed convolution expands a sparse point into an arbitrarily large dense region and multiple regions could overlap with each other (Fig.~\ref{fig:pruning}).

After a transposed convolution, not all the newly created coordinates contain object bounding box anchors. Thus, we remove some of these voxels that have a small probability of containing bounding box anchors. We denote a function that returns the probability given features at each voxel as $P_{\text{s}}(\cdot)$ and remove all voxels $P_\text{s}(\cdot) < \tau$, where $\tau$ is the sparsity pruning confidence threshold.

\begin{align}
\mathbf{p} & = P_{\text{s}}(\mathscr{T}; \mathbf{W}_P) \\
\mathscr{T}' & = \text{SparsityPruning}(\mathscr{T}, \mathbf{p} < \tau)
\end{align}

\subsubsection{Skip Connection and Sparse Tensor Addition}
\label{sec:sparse_tensor_addition}

The upsampled sparse tensor feature maps from the generative process have gone through extreme spatial compression that allows neurons to see larger context, but have lost spatial resolution. To recover the fine details of the input, we create the skip connections to the cached feature map from the encoder~\cite{minkowskinet,FCGF2019}. Since both the upsampled feature map and the lower layer feature map are all sparse tensors, we use sparse tensor addition. This process also expands the support to be the union of the supports of both sparse tensors.

\subsection{Multi-scale Bounding Box Anchor Prediction}
\label{sec:bbox_prediction}

Every voxel after the sparsity pruning potentially contains bounding box anchors. Therefore, we make a direct prediction of the bounding box parameters for every layer of the pruned sparse tensors. Specifically, for each $k$ anchor box, the network predicts 1 object anchor likelihood score, 6 offsets relative to the anchor box, and $c$ semantic class scores. This results in $(c+7)k$ outputs per voxel.

To capture as many shape variations, we use bounding box anchors with different aspect ratios. Specifically, for each anchor ratio seed $a_r$, we use all unique permutations of $\left[\sqrt{a_r}, \sqrt{a_r}, \frac{1}{\sqrt{a_r}}\right]$ as the aspect ratios of an anchor. In total, we use $k=13$ anchors with $a_r \in \{1, 2, 4, \frac{1}{2}, \frac{1}{4}\}$ including the identity ratio.

However, even with these various anchor ratios, it is difficult to capture the extreme scale variation among 3D objects. Thus, we predict anchors at various stages of the decoder to capture the scale variation of 3D objects similar to Liu~\etal~\cite{liu2016ssd}. We construct the anchors at each level to double the size of the anchors at the previous level.


\subsection{Summary of GSDN Feed Forward}

We summarize the feed forward pass of the generative sparse detection networks in Alg.~\ref{alg:ssd-3d}. The algorithm generates $L$ levels of hierarchical sparse tensor feature maps from the previous level feature maps on Line~\ref{alg:gspnet:encoder}. Then, during the generative phase, we extract anchors and associated bounding box information (Line~\ref{alg:gspnet:anchor}), predict sparsity and prune out voxels (Line~\ref{alg:gspnet:pruning}), and apply transposed convolution (Line~\ref{alg:gspnet:convtr}). We add the upsampled sparse tensor to the corresponding sparse tensor feature map from the encoder (Line~\ref{alg:gspnet:add}).

\begin{algorithm}[t]
\DontPrintSemicolon
\KwInput{$\mathscr{T}, f_l(\cdot; \mathbf{W}_l), f^{\text{Tr}}_l(\cdot; \mathbf{W}^\text{Tr}_l), g^b_l(\cdot; \mathbf{W}^b_l), P_\text{s}(\cdot; \mathbf{G}^s_l) \;\text{for}\; l \in [1,..., L], \tau_s$}
\KwOutput{$\left\{\mathbf{B}_l \right\}_l \;\text{for}\; l \in [1,..., L]$}
    $\mathscr{T}_0 \leftarrow \mathscr{T}$\\
    \tcc{Hierarchical Sparse Tensor Encoder \S~\ref{sec:sparse_pyramid}}
    \For{$l \leftarrow 1, ... L$}
    {
        $\mathscr{T}_l \leftarrow f_l(\mathscr{T}_{l - 1})$  \tcp*{Hierarchical feature tensors} \label{alg:gspnet:encoder}
    }
    \tcc{Generative Sparse Tensor Decoder \S~\ref{sec:generative_network}}
    $\mathscr{T}^\text{Tr}_L \leftarrow \mathscr{T}_L$ \\
    \For{$l \leftarrow L, ..., 1$}
    {
        \If{$l < L$}
        {
            $\mathscr{T}^\text{Tr}_l \leftarrow \mathscr{T}^\text{Tr}_l + \mathscr{T}_l$ \tcp*{Skip connection \S \ref{sec:sparse_tensor_addition}} \label{alg:gspnet:add}
        }
        $\mathbf{B}_l \leftarrow g^b_l(\mathscr{T}^\text{Tr}_l)$ \tcp*{Anchor predictions \S \ref{sec:bbox_prediction}} \label{alg:gspnet:anchor}
        $\mathbf{p}_l \leftarrow P^\text{s}_l(\mathscr{T}^\text{Tr}_l)$ \tcp*{Sparsity predictions}
        $\mathscr{T}^\text{Tr}_l \leftarrow \text{SparsityPruning}(\mathscr{T}^\text{Tr}_l, \mathbf{p}_l < \tau)$ \tcp*{Pruning \S \ref{sec:pruning}} \label{alg:gspnet:pruning}
        \If{$l > 1$}
        {
            $\mathscr{T}^{\text{Tr}}_{l + 1} \leftarrow f^{\text{Tr}}_l(\mathscr{T}^{\text{Tr}}_{l})$  \tcp*{Transposed convolution \S \ref{sec:pruning}} \label{alg:gspnet:convtr}
        }
    }
    \Return $\{\mathbf{B}_l\}_l$
\caption{Generative Sparse Detection Networks}\label{alg:ssd-3d}
\end{algorithm}

\subsection{Losses}
The generative sparse detection network has to predict four types of outputs: sparsity prediction, anchor prediction, semantic class, and bounding box regression.
First, the sparsity and anchor prediction are binary classification problems. However, the majority of the predictions are negative as many voxels does not contain positive anchors. Thus, we propose balanced cross entropy loss:
$$
L_\textrm{b}(\hat{\textbf{y}}, \textbf{y}) = - \frac{1}{2|\mathcal{P}|} \sum_{i \in \mathcal{P}} \log(P(\hat{\mathbf{y}}_i)) - \frac{1}{2|\mathcal{N}|}\sum_{i \in \mathcal{N}} \log(1 - P(\hat{\mathbf{y}}_i))
$$
where $\mathcal{P} = \{i | y_i = 1\}$ and $\mathcal{N} = \{i | y_i = 0\}$ are the set of indices with positive and negative labels respectively.
We define an anchor to be positive if any of the anchors in a voxel overlaps with any ground-truth bounding boxes for 3D IoU \textgreater~0.35 and negative if 3D IoU \textless~0.2. As the sparsity prediction must contain all anchors in subsequent levels, we define a sparsity to be positive if any of the subsequent positive anchor associated to the current voxel is positive. We do not enforce loss on anchors that have 0.2~\textless 3D IoU \textless~0.35.



Finally, for positive anchors, we train semantic class prediction of the highest overlapping ground-truth bounding box class with the standard cross entropy, $L_\text{class}$, and bounding box center and size regression parameterized by difference of the center location relative to the size of the anchor and the log difference of the size of the bounding box with the Huber loss~\cite{ren2015faster}, $L_\text{reg}$. The final loss is the weighted sum of all losses:
$$
L = \lambda_{\text{s}}L_{\text{s}}
+ \lambda_{\text{anc}}L_{\text{anc}}
+ \lambda_{\text{class}}L_{\text{class}}
+ \lambda_{\text{reg}}L_{\text{reg}}
$$
where we use $\lambda_{\text{s}}=1$, $\lambda_{\text{anc}}=1$, $\lambda_{\text{class}}=1$, $\lambda_{\text{reg}}=0.1$ for all of our experiments.



\subsection{Prediction post-processing}
We train the network to overestimate the number of bounding box anchors as we label all anchors with 3D IoU \textgreater 0.35 as positives. We filter out overlapping predictions with non-maximum suppression and merge them by computing score-weighted average of all removed bounding boxes to fine tune the final predictions similar to Redmon~\etal~\cite{redmon2017yolo9000}.





\section{Experiments}
\begin{table}[t]
\centering
\resizebox{0.65\textwidth}{!}{
\begin{tabular}{l||c|cc}
\toprule
\textbf{Method}                         & \textbf{Single Shot}    & \textbf{mAP@0.25} & \textbf{mAP@0.5} \\ \midrule
DSS~\cite{DeepSlidingShapes,hou20193d}  & \xmark         & 15.2        & 6.8     \\
MRCNN 2D-3D~\cite{he2017mask,hou20193d} & \xmark         & 17.3        & 10.5    \\
F-PointNet~\cite{qi2018frustum}  & \xmark         & 19.8        & 10.8    \\
GSPN~\cite{yi2019gspn,qi2019deep}       & \xmark         & 30.6        & 17.7    \\
3D-SIS~\cite{hou20193d}                 & \cmark         & 25.4        & 14.6    \\
3D-SIS~\cite{hou20193d} + 5 views       & \cmark         & 40.2        & 22.5    \\
VoteNet~\cite{qi2019deep}               & \xmark         & 58.6        & 33.5    \\ \midrule
\methodname (Ours)                      & \cmark         & \textbf{62.8}        & \textbf{34.8}  \\ \bottomrule
\end{tabular}
}
\vspace{1em}
\caption{Object detection mAP on the ScanNet v2 validation set. DSS, MRCNN 2D-3D, FPointNet are from~\cite{hou20193d}. GSPN from~\cite{qi2019deep}. Our method, despite being single-shot, outperforms all previous state-of-the-art methods.}
\label{tab:scannet}
\vspace{-2em}
\end{table}
\begin{table}[t]
\centering
\resizebox{\textwidth}{!}{
\begin{tabular}{l||cccccccccccccccccc|c}
\toprule
              & cab & bed & chair & sofa & tabl & door & wind & bkshf & pic & cntr & desk & curt & fridg & showr & toil & sink & bath & ofurn & mAP \\ \midrule
Hou~\etal~\cite{hou20193d} & 12.75 & 63.14 & 65.98 & 46.33 & 26.91 & 7.95 & 2.79 & 2.30 & 0.00 & 6.92 & 33.34 & 2.47 & 10.42 & 12.17 & 74.51 & 22.87 & 58.66 & 7.05 & 25.36 \\
Hou~\etal~\cite{hou20193d} + 5 views & 19.76 & 69.71 & 66.15 & 71.81 & 36.06 & 30.64 & 10.88 & 27.34 & 0.00 & 10.00 & 46.93 & 14.06 & \textbf{53.76} & 35.96 & 87.60 & 42.98 & 84.30 & 16.20 & 40.23 \\
Qi~\etal~\cite{qi2019deep}  & 36.27 & \textbf{87.92} & 88.71 & \textbf{89.62} & 58.77 & \textbf{47.32} & 38.10 & 44.62 & 7.83 & 56.13 & \textbf{71.69} & 47.23 & 45.37 & 57.13 & 94.94 & 54.70 & 92.11 & 37.20 & 58.65 \\ \midrule
\methodname (Ours) & \textbf{41.58} & 82.50 & \textbf{92.14} & 86.95 & \textbf{61.05} & 42.41 & \textbf{40.66} & \textbf{51.14} & \textbf{10.23} & \textbf{64.18} & 71.06 & \textbf{54.92} & 40.00 & \textbf{70.54} & \textbf{99.97} & \textbf{75.50} & \textbf{93.23} & \textbf{53.07} & \textbf{62.84} \\ \bottomrule
\end{tabular}}
\vspace{1em}
\caption{Class-wise mAP@0.25 object detection result on the ScanNet v2 validation set. Our method outperforms previous state-of-the-art on majority of the semantic classes.}
\label{tab:scannet_25}
\vspace{-1em}
\end{table}

We evaluate our method on three 3D indoor datasets and compare with state-of-the-art object detection methods (\ref{sec:comparison}). We also make a detailed analysis of the speed and memory footprint of our method (\ref{sec:analysis}). Finally, we demonstrate the scalability of our proposed method on extremely large scenes (\ref{sec:scalability}). 

\noindent\textbf{Datasets.} We evaluate our method on the ScanNet dataset~\cite{dai2017scannet}, annotated 3D reconstructions of 1500 indoor scenes with instance labels of 18 semantic classes.
We follow the experiment protocol of Qi~\etal~\cite{qi2019deep} to define axis-aligned bounding boxes that encloses all points of an instance without any margin as the ground truth bounding boxes.

The second dataset is the Stanford Large-Scale 3D Indoor Spaces (S3DIS) dataset~\cite{armeni_cvpr16}. It contains 3D scans of 6 buildings with 272 rooms, each with instance and semantic labels of 7 structural elements such as floor and ceiling, and five furniture classes. We train and evaluate our method on the official furniture split and use the most-widely used \textit{Area 5} for our test split. We follow the same procedure as above to generate ground-truth bounding boxes from instance labels.

Finally, we demonstrate the scalability of \methodname on the Gibson environment~\cite{xiazamirhe2018gibsonenv} as it contains high-quality reconstructions of 575 multi-story buildings. 

\noindent\textbf{Metrics.} We adopt the average precision (AP) and class-wise mean AP (mAP) to evaluate the performance of object detectors following the widely used convention of 2D object detection. We consider a detection as a positive match when a 3D intersection-over-union(IoU) between the prediction and the ground-truth bounding box is above a certain threshold.


\noindent\textbf{Training hyper-parameters.} We train our models using SGD optimizer with exponential decay of learning rate from 0.1 to 1e-3 for 120k iterations with the batch size 16. As our model can process an entire scene fully-convolutionally, we do not make smaller crops of a scene. We use high-dimensional ResNet34~\cite{minkowskinet,he2016deep} for the encoder. For all experiments, we use voxel size of 5cm, transpose kernel size of 3, with $L=4$ scale hierarchy, sparsity pruning confidence $\tau=0.3$, and 3D NMS threshold 0.2.

\subsection{Object detection performance analysis}
\label{sec:comparison}

\begin{figure}[t]
    \centering
    \includegraphics[width=0.393\textwidth]{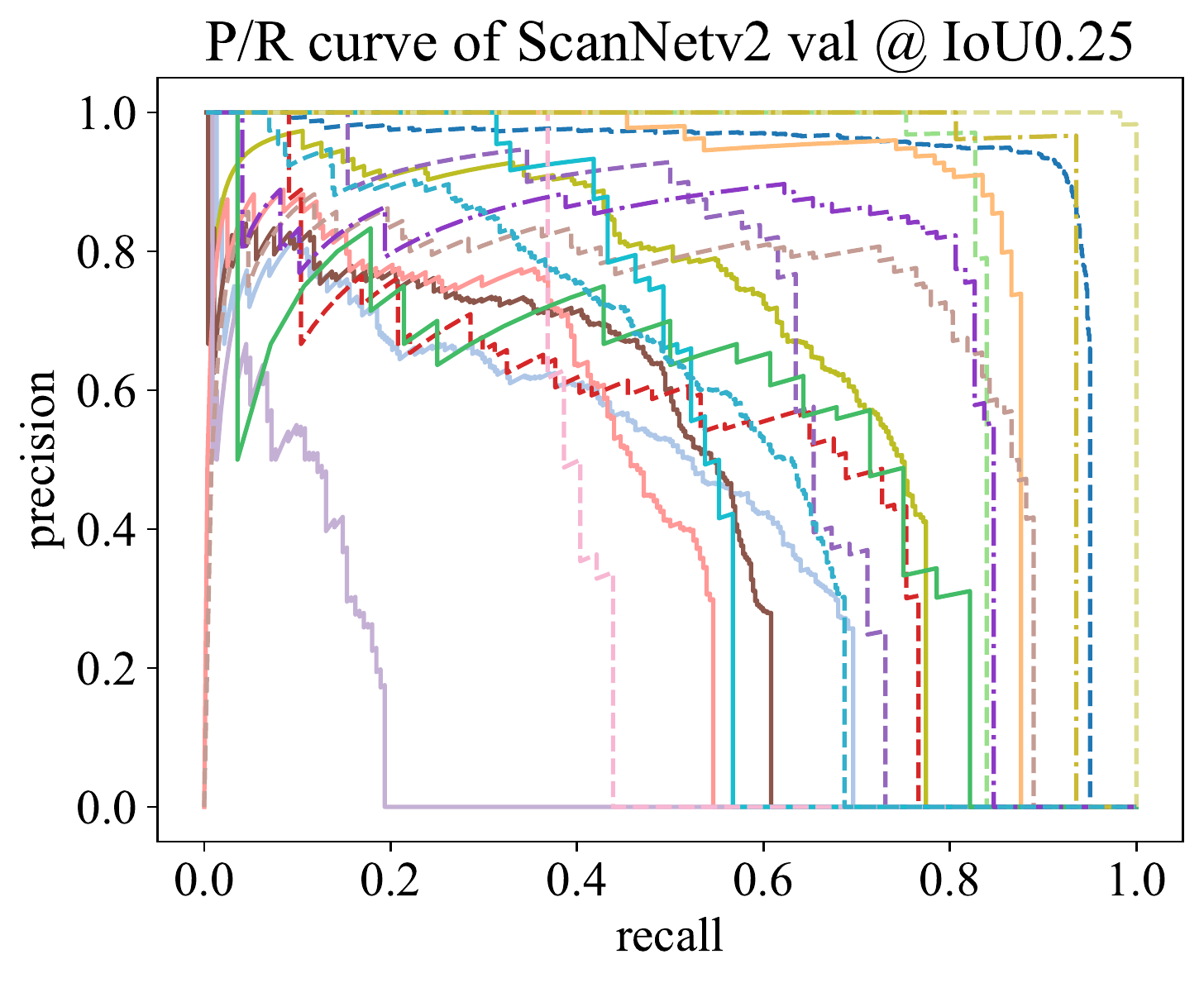} \includegraphics[width=0.507\textwidth]{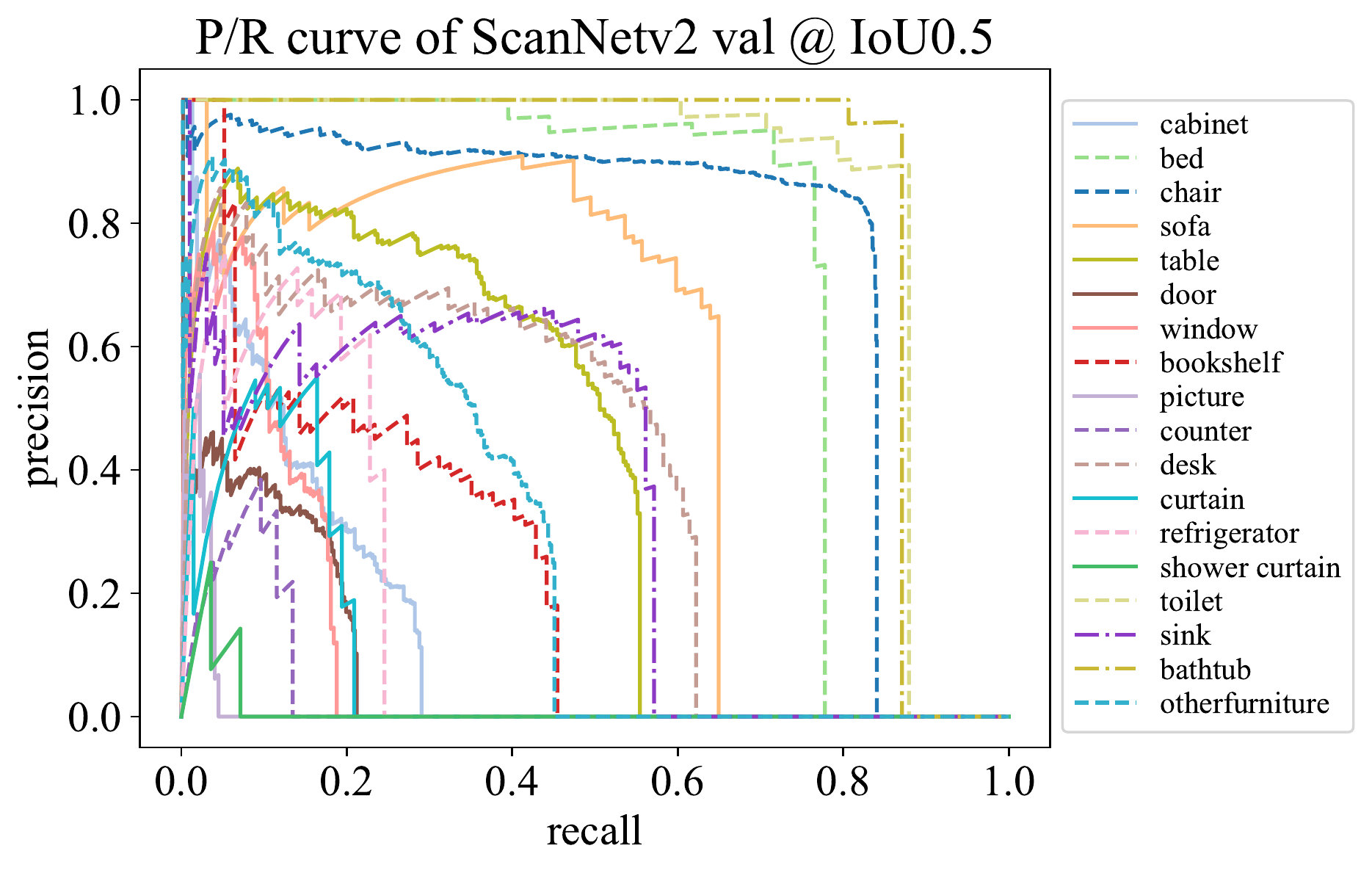}
    \vspace{-0.5em}
    \caption{Per-class precision/recall curve of ScanNetV2 validation object detection.}
    \label{fig:scannet_prcurve}
\end{figure}
\begin{table}[t]
\centering
\setlength{\tabcolsep}{2pt}
\resizebox{0.7\textwidth}{!}{
\begin{tabular}{c|c|c||ccccc|c}
\toprule
IoU Thres. & Metric & Method & table & chair & sofa & bookcase & board & avg \\ \midrule
\multirow{4}{*}{0.25}&\multirow{2}{*}{AP}    & Yang~\etal~\cite{yang2019learning}* & 27.33 & 53.41 & 9.09 & 14.76 & 29.17 & 26.75 \\ 
                             && \methodname (ours)     & 73.69 & 98.11 & 20.78 & 33.38 & 12.91 & 47.77        \\ \cmidrule{2-9}
&\multirow{2}{*}{Recall} & Yang~\etal~\cite{yang2019learning}* & 40.91 & 68.22 & 9.09 & 29.03 & 50.00 & 39.45 \\ 
                              && \methodname (ours)     & 85.71 & 98.84 & 36.36 & 61.57 & 26.19 & 61.74        \\ \midrule
\multirow{4}{*}{0.5}&\multirow{2}{*}{AP}    & Yang~\etal~\cite{yang2019learning}* & 4.02 & 17.36 & 0.0 & 2.60 & 13.57 & 7.51 \\ 
                         &     & \methodname (ours)     & 36.57 & 75.29 & 6.06  & 6.46  & 1.19 & 25.11         \\ \cmidrule{2-9}
&\multirow{2}{*}{Recall} & Yang~\etal~\cite{yang2019learning}* & 16.23 & 38.37 & 0.0 & 12.44 & 33.33 & 20.08 \\ 
 &                             & \methodname (ours)  & 50.00 & 82.56 & 18.18 & 18.52 & 2.38 & 34.33         \\ \bottomrule
\end{tabular}
}
\vspace{1em}
\caption{Object detection result on furniture subclass of S3DIS dataset building 5. *: Converted the instance segmentation results to bounding boxes for reference}
\label{tab:s3dis}
\vspace{-2em}
\end{table}

We compare the object detection performance of our proposed method with the previous state-of-the-art methods on Table~\ref{tab:scannet} and Table~\ref{tab:scannet_25}. Our method, despite being a single-shot detector, outperforms all two-stage baselines with 4.2\% mAP@0.25 and 1.3\% mAP@0.5 performance gain and outperforms the state-of-the-art on the majority of semantic classes.

We also report the S3DIS detection results on Table~\ref{tab:s3dis}. 
It is also worth noting that Yang~\etal~\cite{yang2019learning} and some preceding works~\cite{wang2018sgpn,wang2019associatively} crop a scene into multiple 1m$\times$1m floor areas, and merge them with heuristics~\cite{wang2018sgpn}, which not only heavily restricts the receptive field but also require slow pre-processing and post-processing. Our method in contrast takes the whole scene as an input.

We plot class-wise precision-recall curves of ScanNet validation set on Figure~\ref{fig:scannet_prcurve}. We found that some of the PR curves drop sharply, which indicates that the simple aspect-ratio anchors have a low recall.

\begin{figure}[htp!]
\centering
\begin{tabular}{cccc}
G.T. & Hou~\etal~\cite{hou20193d} & Qi~\etal~\cite{qi2019deep} & Ours \\
\includegraphics[width=.24\textwidth]{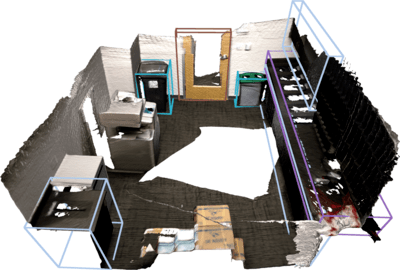} &
\includegraphics[width=.24\textwidth]{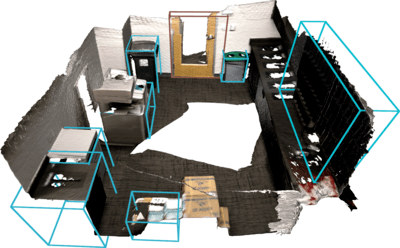} &
\includegraphics[width=.24\textwidth]{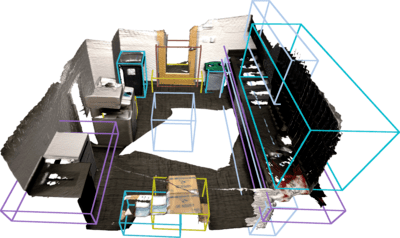} &
\includegraphics[width=.24\textwidth]{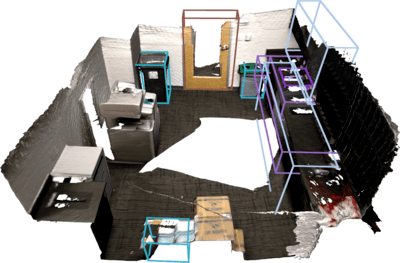} \\
\includegraphics[width=.24\textwidth]{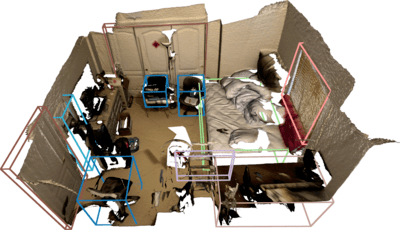} &
\includegraphics[width=.24\textwidth]{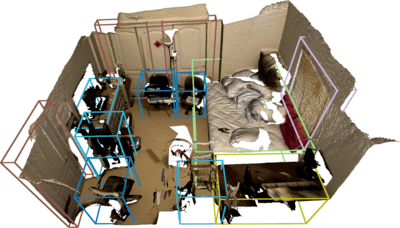} &
\includegraphics[width=.24\textwidth]{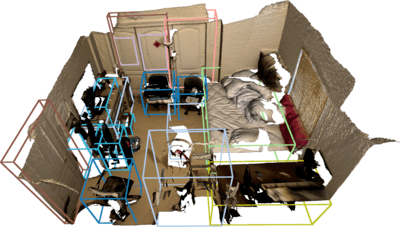} &
\includegraphics[width=.24\textwidth]{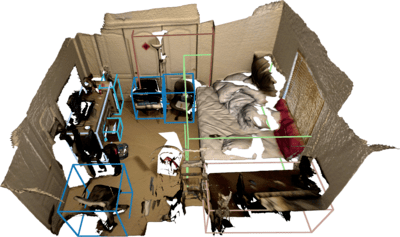} \\
\includegraphics[width=.24\textwidth]{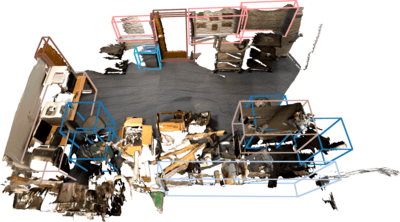} &
\includegraphics[width=.24\textwidth]{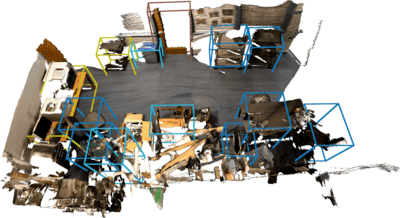} &
\includegraphics[width=.24\textwidth]{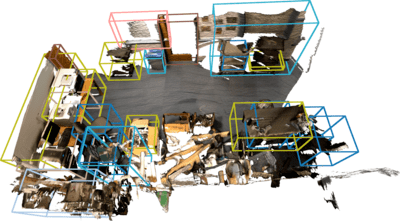} &
\includegraphics[width=.24\textwidth]{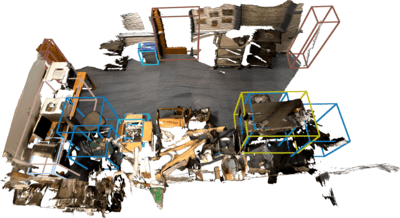} \\
\includegraphics[width=.24\textwidth]{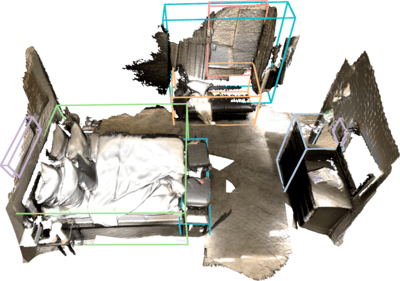} &
\includegraphics[width=.24\textwidth]{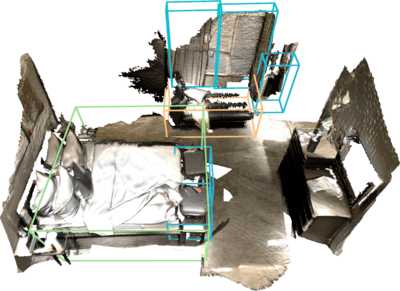} &
\includegraphics[width=.24\textwidth]{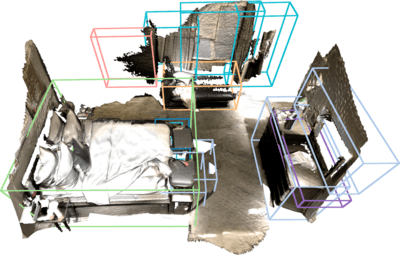} &
\includegraphics[width=.24\textwidth]{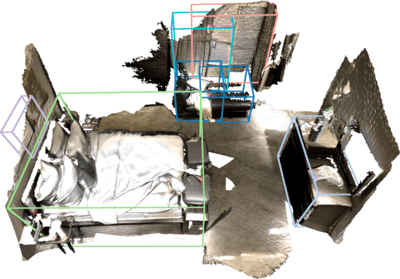} \\
\includegraphics[width=.24\textwidth]{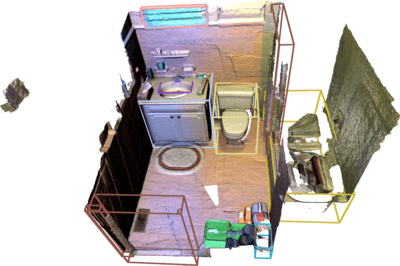} &
\includegraphics[width=.24\textwidth]{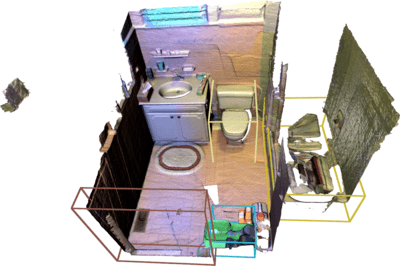} &
\includegraphics[width=.24\textwidth]{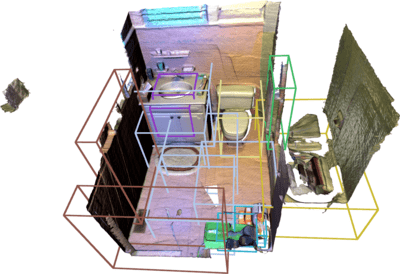} &
\includegraphics[width=.24\textwidth]{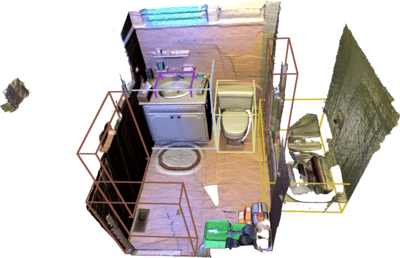} \\
\end{tabular}
\caption{Qualitative object detection results on the ScanNet dataset.}
\label{fig:scannet_results}
\vspace{-2em}
\end{figure}
\begin{figure}[htp!]
\centering
\begin{tabular}{cccc}
G.T. & Ours & G.T. & Ours \\
\includegraphics[width=.2\textwidth]{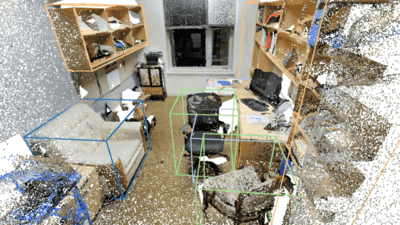} &
\includegraphics[width=.2\textwidth]{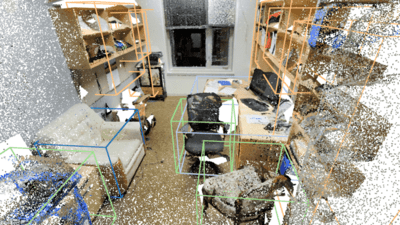} &
\includegraphics[width=.2\textwidth]{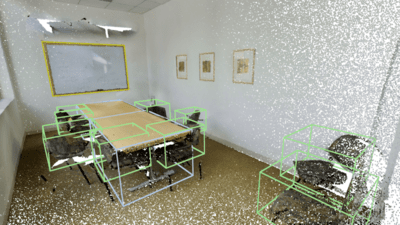} &
\includegraphics[width=.2\textwidth]{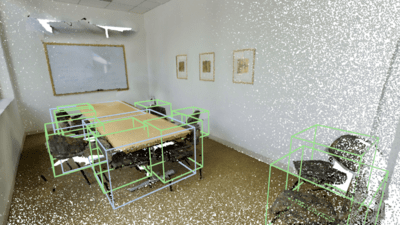} \\
\includegraphics[width=.2\textwidth]{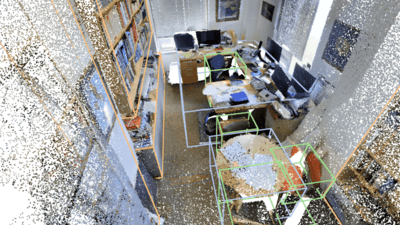} &
\includegraphics[width=.2\textwidth]{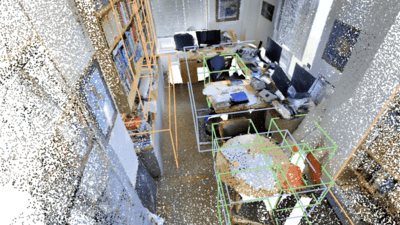} &
\includegraphics[width=.2\textwidth]{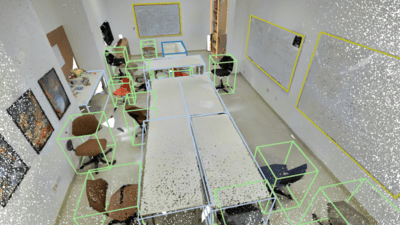} &
\includegraphics[width=.2\textwidth]{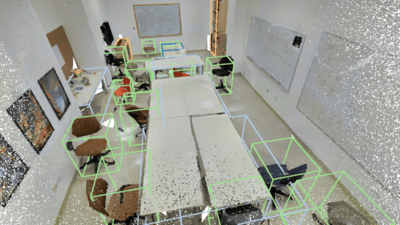} \\
\includegraphics[width=.2\textwidth]{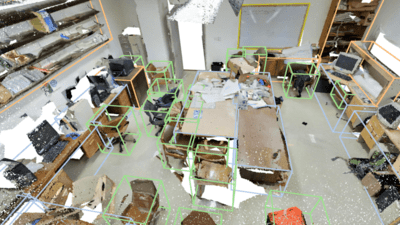} &
\includegraphics[width=.2\textwidth]{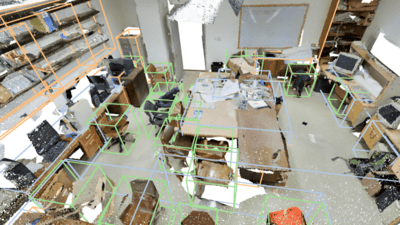} &
\includegraphics[width=.2\textwidth]{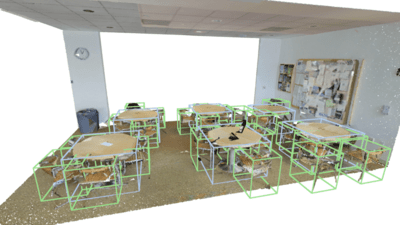} &
\includegraphics[width=.2\textwidth]{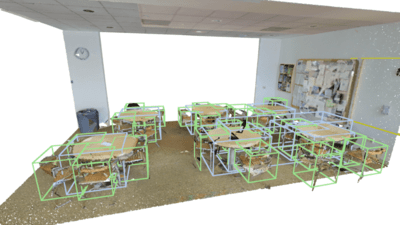} \\
\includegraphics[width=.2\textwidth]{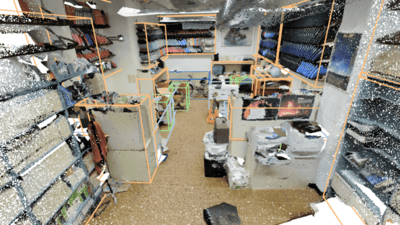} &
\includegraphics[width=.2\textwidth]{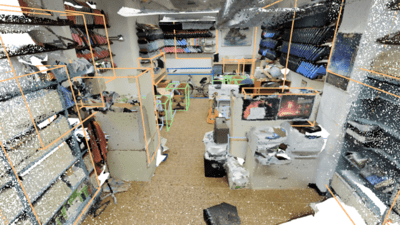} &
\includegraphics[width=.2\textwidth]{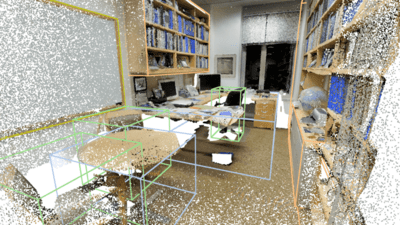} &
\includegraphics[width=.2\textwidth]{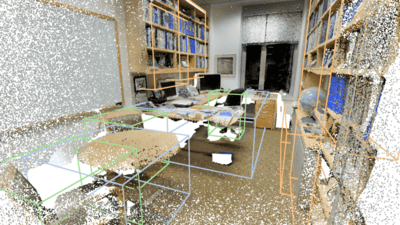} \\
\end{tabular}
\vspace{-0.5em}
\caption{Qualitative object detection results on the S3DIS dataset.}
\vspace{-0.5em}
\label{fig:stanford_results}
\end{figure}

Finally, we visualize qualitative results of our method on Figure~\ref{fig:scannet_results} and Figure~\ref{fig:stanford_results}. In general, we found that our method suffers from detecting thin structures such as bookcase and board, which may be resolved by adding more extreme-shaped anchors. 
Please refer to the supplementary materials for the class-wise breakdown of mAP@0.5 on the ScanNet dataset and class-wise precision-recall curves for the S3DIS dataset.


\begin{figure}[ht!]
\centering
  \includegraphics[width=0.45\linewidth]{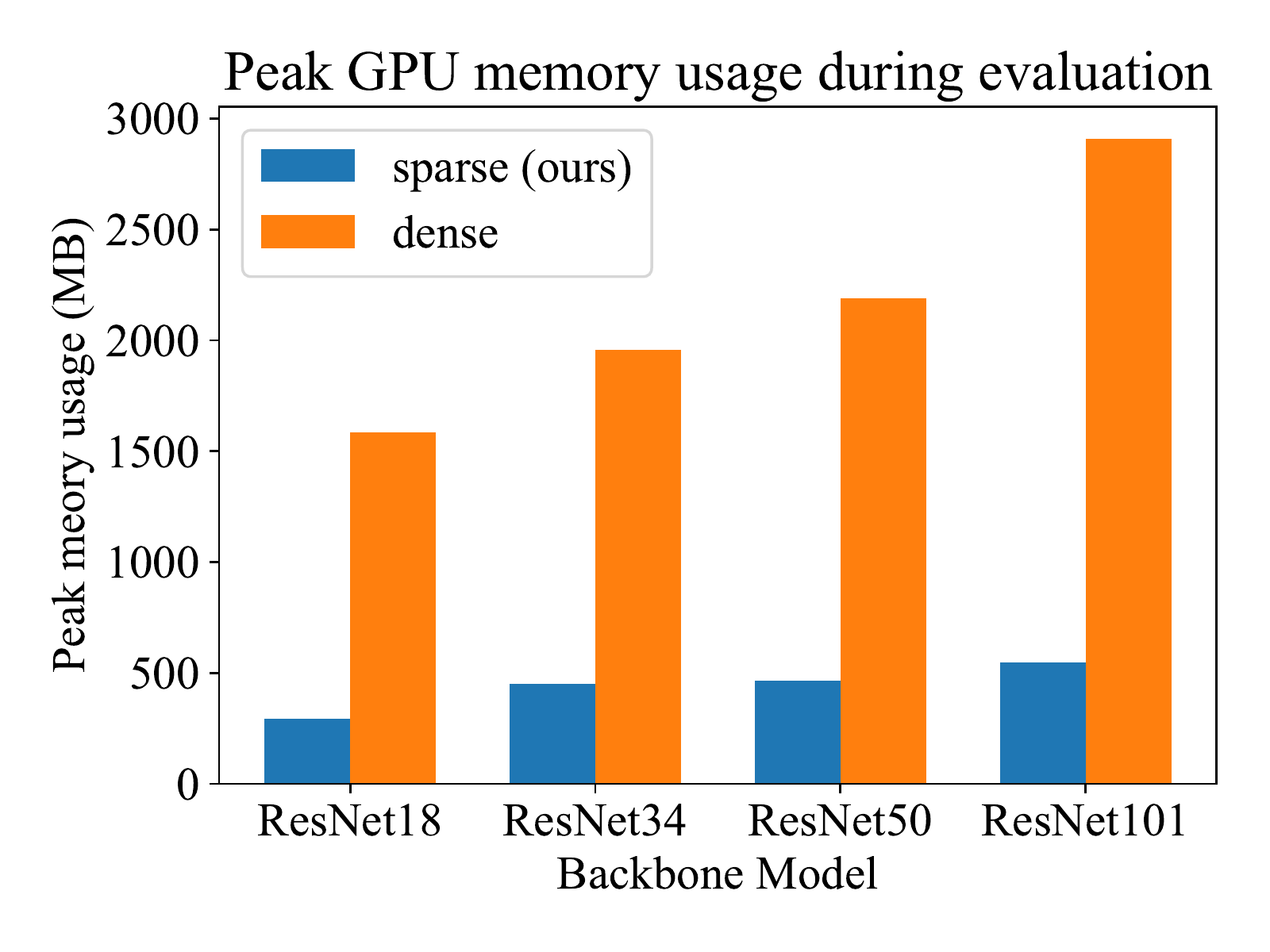}
  \includegraphics[width=0.45\linewidth]{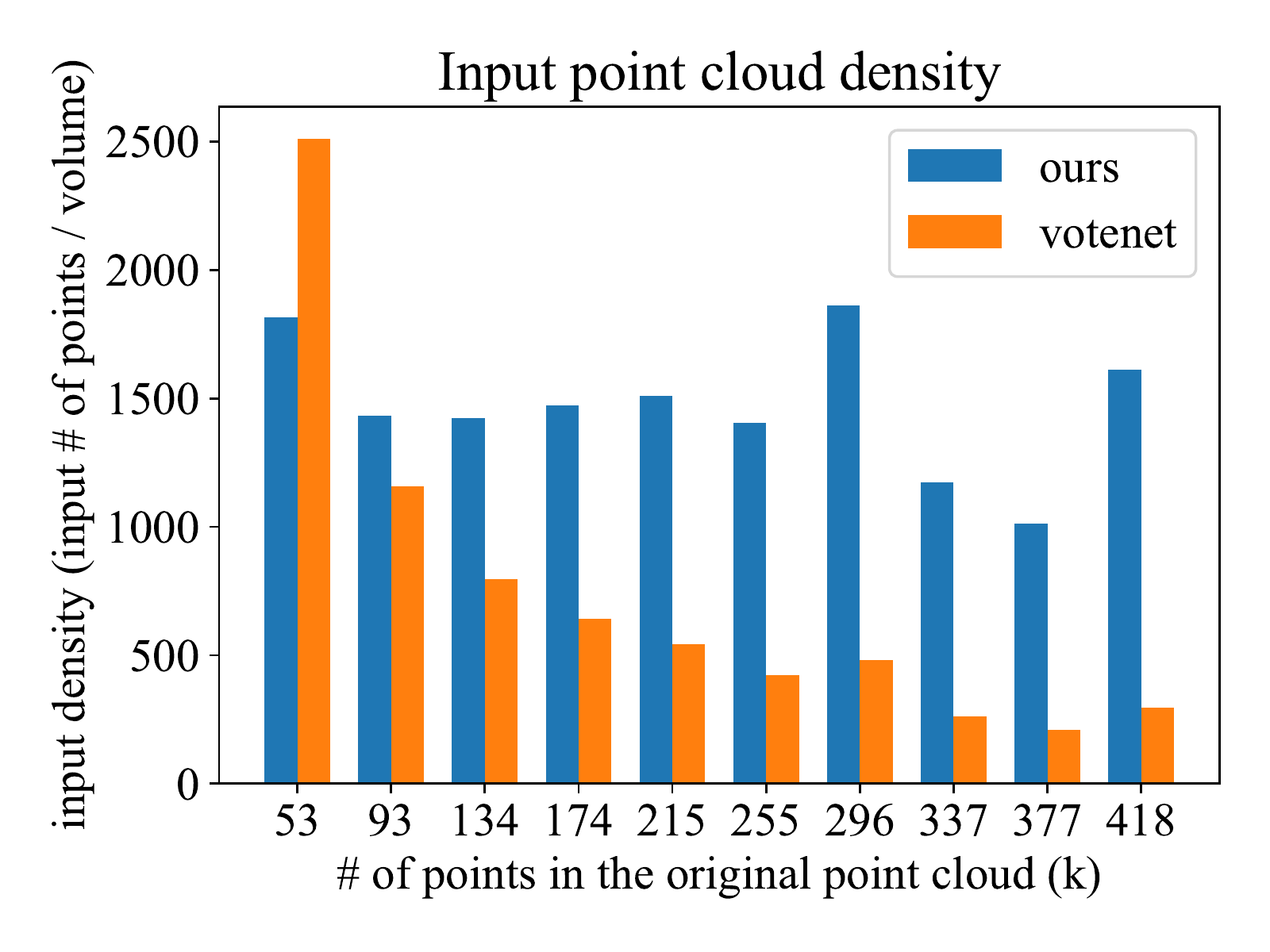}
  \vspace{-0.5em}
  \caption{Left: Memory usage comparison on ScanNet dataset evaluation: Our proposed sparse encoder and decoder maintains low memory usage compared to the dense counterparts. Right: Point cloud density on ScanNet dataset. Our model maintains constant input point cloud density compared to Qi~\etal~\cite{qi2019deep}, which samples constant number of points regardless of the size of the input.}
  \label{fig:memory_and_density}
  \vspace{-1em}
\end{figure}

\begin{figure}[t]
    \centering
    \includegraphics[width=0.45\textwidth]{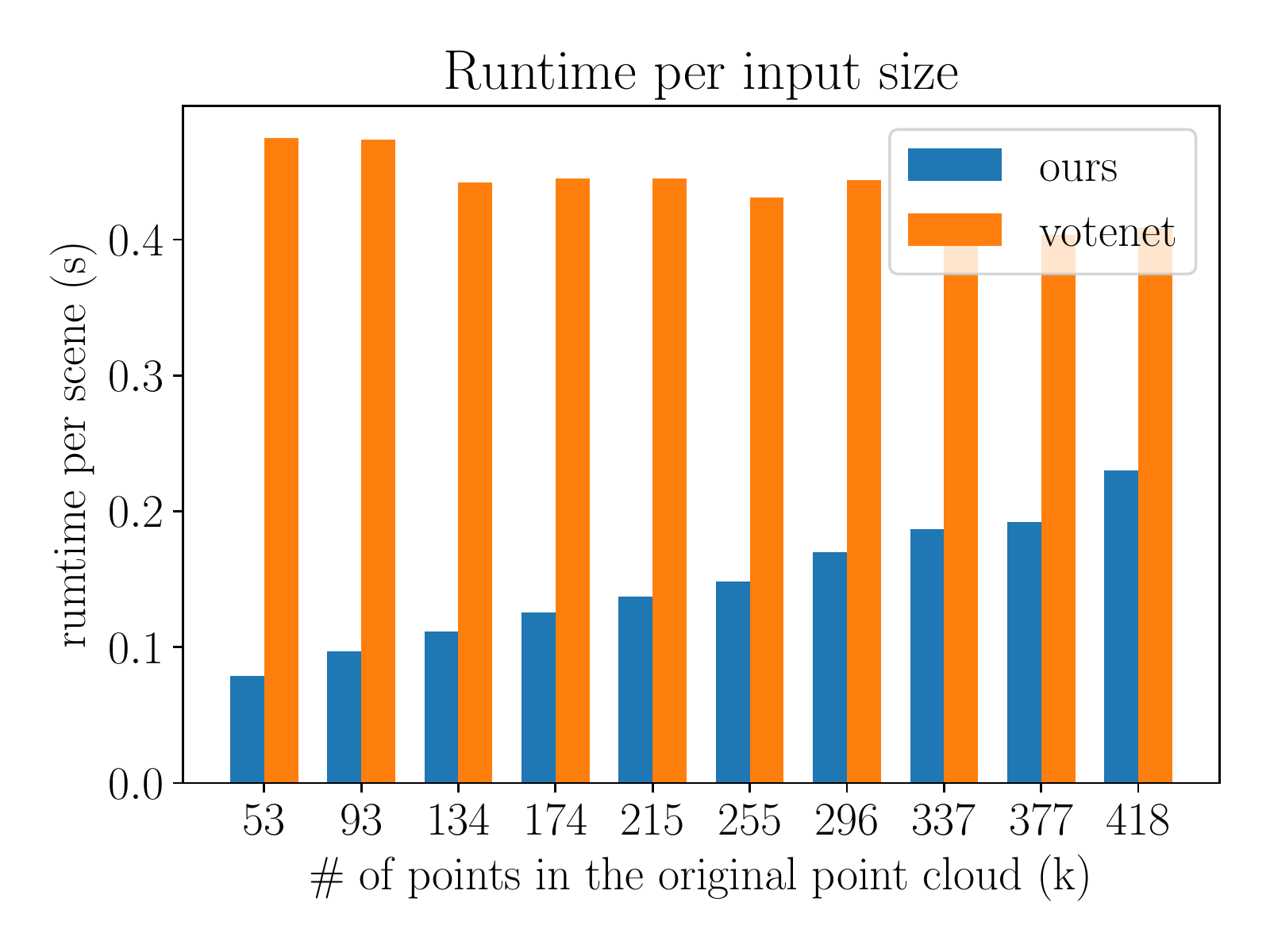}
    \includegraphics[width=0.45\textwidth]{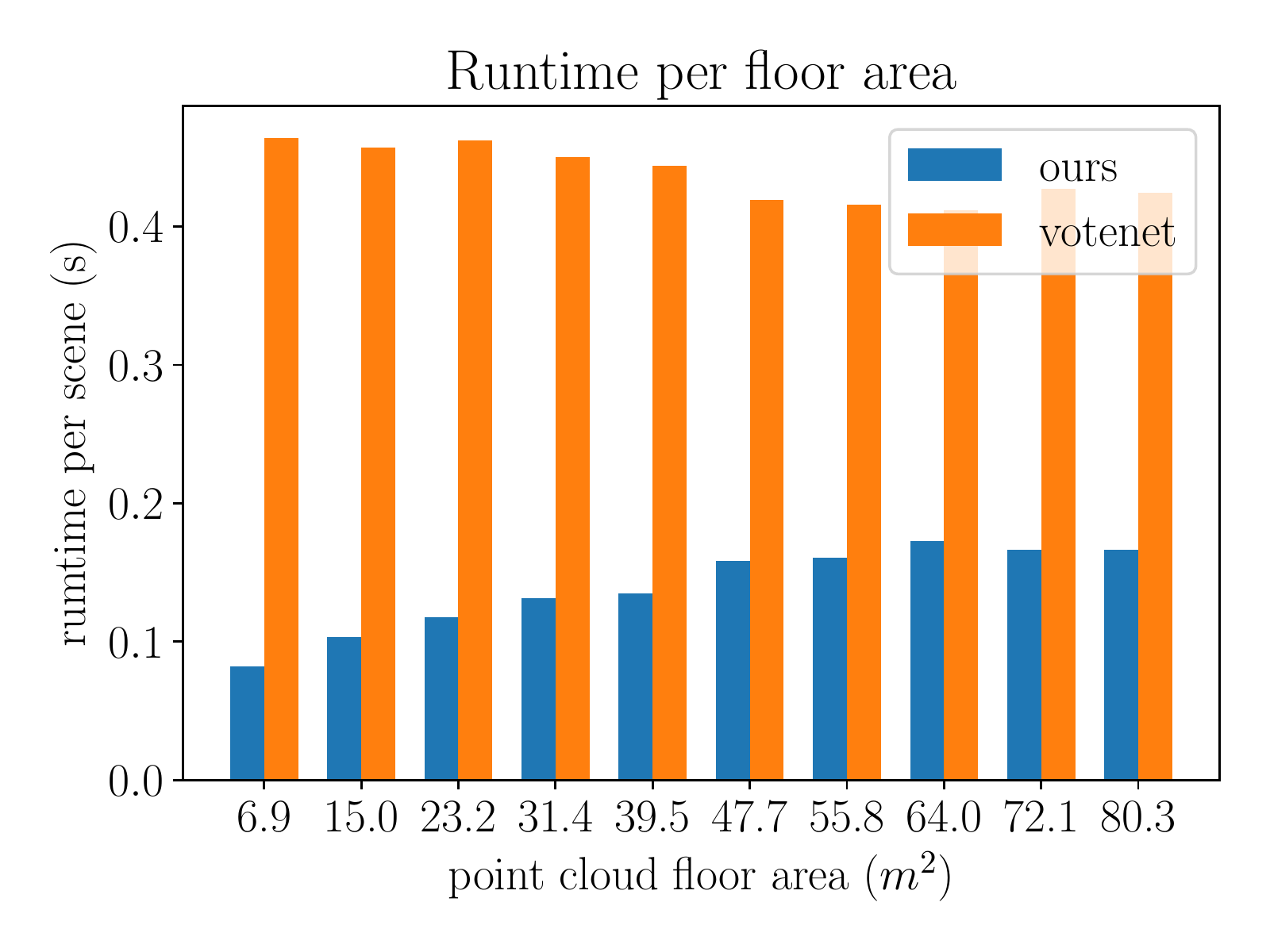}
    \vspace{-0.5em}
    \caption{Runtime comparison on ScanNet v2 validation set: Qi~\etal~\cite{qi2019deep} samples a constant number of points from a scene and their post-processing is inversely proportional to the density, whereas our method scales linearly to the number of points, and sublinearly to the floor area while being significantly faster.}
    \label{fig:scannet_runtime}
\end{figure}
\subsection{Speed and Memory Analysis}
\label{sec:analysis}

We analyze the memory footprint and runtime in Figure~\ref{fig:memory_and_density} and Figure~\ref{fig:scannet_runtime}. For the memory analysis, we compare our method with the dense object detector~\cite{hou20193d} and measured the peak memory usage on ScanNetV2 validation set. As expected, our proposed network maintains extremely low memory consumption regardless of the depth of the network while that of the dense counterparts grows noticeably.

For runtime analysis, we compare the network feed forward and post-processing time of our method with Qi~\etal~\cite{qi2019deep} in Figure~\ref{fig:scannet_runtime}. On average, our method takes 0.12 seconds while Qi~\etal~\cite{qi2019deep} takes 0.45 seconds to process a scene of ScanNetV2 validation set. Moreover, the runtime of our method grows linearly to the number of points and sublinearly to the floor area of the point cloud, due to the sparsity of our point representation.
Note that Qi~\etal~\cite{qi2019deep} subsamples a constant number of points from input point clouds regardless of the size of the input point clouds.
Thus, the point density of Qi~\etal~\cite{qi2019deep} changes significantly as the point cloud gets larger.
However, our method maintains the constant density as shown in Figure~\ref{fig:memory_and_density}, which allows our method to scale to extremely large scenes as shown in Section~\ref{sec:scalability}.
In sum, we achieve $3.78\times$ speed up and 4.2\% mAP@0.25 performance gain compared to Qi~\etal~\cite{qi2019deep} while maintaining the same point density from small to large scenes.
 

\subsection{Scalability and generalization of \methodname on extremely large inputs}
\label{sec:scalability}

We qualitatively demonstrate the scalability and generalization ability of our method on large scenes from the S3DIS dataset~\cite{armeni_cvpr16} and the Gibson environment~\cite{xiazamirhe2018gibsonenv}.
First, we process the entire building 5 of S3DIS which consists of 78M points, 13984m${}^3$ volume, and 53 rooms.
\methodname takes 20 seconds for a single feed-forward of the entire scene including data pre-processing and post-processing. The model uses 5G GPU memory to detect 573 instances of 3D objects, which we visualized on Figure~\ref{fig:stanford_full}.

\begin{figure}[h]
    \centering
    \includegraphics[width=0.45\textwidth]{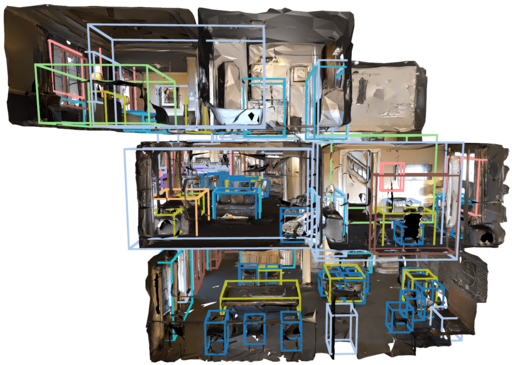}
    \includegraphics[width=0.52\textwidth]{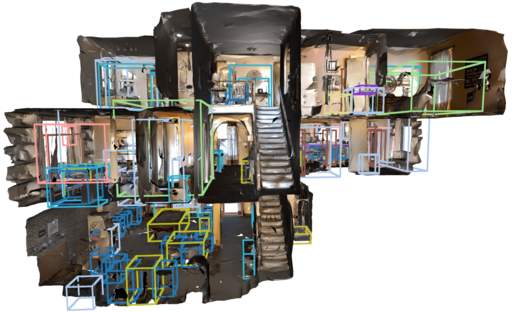}
    \caption{Detection results on a Gibson environment scene \textit{Uvalda}~\cite{xiazamirhe2018gibsonenv}: \methodname can process a 17-room building with 1.4M points in a single \textit{fully-convolutional} feed-forward pass.} \label{fig:gibson}
    \vspace{-1em}
\end{figure}

Similarly, we train our network on ScanNet dataset~\cite{dai2017scannet} which only contain single-floor 3D scans. However, we tested the network on multi-story buildings. On Figure~\ref{fig:gibson}, we visualize our detection results on the scene named \textit{Uvalda} from Gibson, which is a 3-story building with 173m${}^2$ floor area. Note that our fully-convolutional network, which was only trained on single-story 3D scans, generalizes to multi-story buildings without any ad-hoc pre-processing or post-processing.
\methodname takes 2.2 seconds to process the building from the raw point cloud and takes up 1.8G GPU memory to detect 129 instances of 3D objects.

\section{Conclusion}

In this work, we present the Generative Sparse Detection Network (\methodname) for single-shot fully-convolutional 3D object detection. \methodname maintains sparsity throughout the network by generating object centers using the proposed generative sparse tensor decoder. \methodname can efficiently process large-scale point clouds without cropping the scene into smaller windows to take advantage of the full receptive field. Thus, \methodname outperforms the previous state-of-the-art method by 4.2 mAP@0.25 while being $3.78\times$ faster. In the follow-up work, we will examine and adopt various image detection techniques to boost the accuracy of \methodname.

%
%
\nocite{tange2011gnu}
\bibliographystyle{splncs04}
\bibliography{egbib}

\begin{thebibliography}{10}
\providecommand{\url}[1]{\texttt{#1}}
\providecommand{\urlprefix}{URL }
\providecommand{\doi}[1]{https://doi.org/#1}

\bibitem{armeni_cvpr16}
Armeni, I., Sener, O., Zamir, A.R., Jiang, H., Brilakis, I., Fischer, M.,
  Savarese, S.: 3d semantic parsing of large-scale indoor spaces. In:
  Proceedings of the IEEE International Conference on Computer Vision and
  Pattern Recognition (2016)

\bibitem{minkowskinet}
Choy, C., Gwak, J., Savarese, S.: 4d spatio-temporal convnets: Minkowski
  convolutional neural networks. In: Proceedings of the IEEE Conference on
  Computer Vision and Pattern Recognition. pp. 3075--3084 (2019)

\bibitem{FCGF2019}
Choy, C., Park, J., Koltun, V.: Fully convolutional geometric features. In:
  ICCV (2019)

\bibitem{choy20163d}
Choy, C.B., Xu, D., Gwak, J., Chen, K., Savarese, S.: 3d-r2n2: A unified
  approach for single and multi-view 3d object reconstruction. In: Proceedings
  of the European Conference on Computer Vision ({ECCV}) (2016)

\bibitem{dai2017scannet}
Dai, A., Chang, A.X., Savva, M., Halber, M., Funkhouser, T., Nie{\ss}ner, M.:
  Scannet: Richly-annotated 3d reconstructions of indoor scenes. In:
  Proceedings of the IEEE Conference on Computer Vision and Pattern
  Recognition. pp. 5828--5839 (2017)

\bibitem{dai2019sg}
Dai, A., Diller, C., Nie{\ss}ner, M.: Sg-nn: Sparse generative neural networks
  for self-supervised scene completion of rgb-d scans. arXiv preprint
  arXiv:1912.00036  (2019)

\bibitem{dai2018scancomplete}
Dai, A., Ritchie, D., Bokeloh, M., Reed, S., Sturm, J., Nie{\ss}ner, M.:
  Scancomplete: Large-scale scene completion and semantic segmentation for 3d
  scans. In: Proc. Computer Vision and Pattern Recognition (CVPR), IEEE (2018)

\bibitem{sparseconvnet}
Graham, B., Engelcke, M., van~der Maaten, L.: 3{D} semantic segmentation with
  submanifold sparse convolutional networks. CVPR  (2018)

\bibitem{SubmanifoldSparseConvNet}
Graham, B., van~der Maaten, L.: Submanifold sparse convolutional networks.
  arXiv preprint arXiv:1706.01307  (2017)

\bibitem{han2015deep}
Han, S., Mao, H., Dally, W.J.: Deep compression: Compressing deep neural
  networks with pruning, trained quantization and huffman coding. arXiv
  preprint arXiv:1510.00149  (2015)

\bibitem{he2017mask}
He, K., Gkioxari, G., Doll{\'a}r, P., Girshick, R.: Mask r-cnn. In: Proceedings
  of the IEEE international conference on computer vision. pp. 2961--2969
  (2017)

\bibitem{he2016deep}
He, K., Zhang, X., Ren, S., Sun, J.: Deep residual learning for image
  recognition. In: Proceedings of the IEEE conference on computer vision and
  pattern recognition. pp. 770--778 (2016)

\bibitem{hou20193d}
Hou, J., Dai, A., Nie{\ss}ner, M.: 3d-sis: 3d semantic instance segmentation of
  rgb-d scans. In: Proceedings of the IEEE Conference on Computer Vision and
  Pattern Recognition. pp. 4421--4430 (2019)

\bibitem{lahoud20193d}
Lahoud, J., Ghanem, B., Pollefeys, M., Oswald, M.R.: 3d instance segmentation
  via multi-task metric learning. In: Proceedings of the IEEE International
  Conference on Computer Vision. pp. 9256--9266 (2019)

\bibitem{li2016vehicle}
Li, B., Zhang, T., Xia, T.: Vehicle detection from 3d lidar using fully
  convolutional network. arXiv preprint arXiv:1608.07916  (2016)

\bibitem{lin2017feature}
Lin, T.Y., Doll{\'a}r, P., Girshick, R., He, K., Hariharan, B., Belongie, S.:
  Feature pyramid networks for object detection. In: Proceedings of the IEEE
  conference on computer vision and pattern recognition. pp. 2117--2125 (2017)

\bibitem{liu2019masc}
Liu, C., Furukawa, Y.: Masc: multi-scale affinity with sparse convolution for
  3d instance segmentation. arXiv preprint arXiv:1902.04478  (2019)

\bibitem{liu2016ssd}
Liu, W., Anguelov, D., Erhan, D., Szegedy, C., Reed, S., Fu, C.Y., Berg, A.C.:
  Ssd: Single shot multibox detector. In: European conference on computer
  vision. pp. 21--37. Springer (2016)

\bibitem{maturana_iros_2015}
Maturana, D., Scherer, S.: {VoxNet: A 3D Convolutional Neural Network for
  Real-Time Object Recognition}. In: {IROS} (2015)

\bibitem{Meschedar2019}
Mescheder, L., Oechsle, M., Niemeyer, M., Nowozin, S., Geiger, A.: Occupancy
  networks: Learning 3d reconstruction in function space. In: Proceedings IEEE
  Conf. on Computer Vision and Pattern Recognition (CVPR) (2019)

\bibitem{narang2017exploring}
Narang, S., Elsen, E., Diamos, G., Sengupta, S.: Exploring sparsity in
  recurrent neural networks. arXiv preprint arXiv:1704.05119  (2017)

\bibitem{parashar2017scnn}
Parashar, A., Rhu, M., Mukkara, A., Puglielli, A., Venkatesan, R., Khailany,
  B., Emer, J., Keckler, S.W., Dally, W.J.: Scnn: An accelerator for
  compressed-sparse convolutional neural networks. ACM SIGARCH Computer
  Architecture News  \textbf{45}(2),  27--40 (2017)

\bibitem{Park_2019_CVPR}
Park, J.J., Florence, P., Straub, J., Newcombe, R., Lovegrove, S.: Deepsdf:
  Learning continuous signed distance functions for shape representation. In:
  The IEEE Conference on Computer Vision and Pattern Recognition (CVPR) (June
  2019)

\bibitem{qi2019deep}
Qi, C.R., Litany, O., He, K., Guibas, L.J.: Deep hough voting for 3d object
  detection in point clouds. In: Proceedings of the IEEE International
  Conference on Computer Vision. pp. 9277--9286 (2019)

\bibitem{qi2018frustum}
Qi, C.R., Liu, W., Wu, C., Su, H., Guibas, L.J.: Frustum pointnets for 3d
  object detection from rgb-d data. In: Proceedings of the IEEE Conference on
  Computer Vision and Pattern Recognition. pp. 918--927 (2018)

\bibitem{redmon2017yolo9000}
Redmon, J., Farhadi, A.: Yolo9000: better, faster, stronger. In: Proceedings of
  the IEEE conference on computer vision and pattern recognition. pp.
  7263--7271 (2017)

\bibitem{ren2015faster}
Ren, S., He, K., Girshick, R., Sun, J.: Faster r-cnn: Towards real-time object
  detection with region proposal networks. In: Advances in neural information
  processing systems. pp. 91--99 (2015)

\bibitem{DeepSlidingShapes}
Song, S., Xiao, J.: {D}eep {S}liding {S}hapes for amodal 3{D} object detection
  in {RGB-D} images. In: CVPR (2016)

\bibitem{tange2011gnu}
Tange, O., et~al.: Gnu parallel-the command-line power tool. The USENIX
  Magazine  \textbf{36}(1),  42--47 (2011)

\bibitem{ogn2017}
Tatarchenko, M., Dosovitskiy, A., Brox, T.: Octree generating networks:
  Efficient convolutional architectures for high-resolution 3d outputs. In:
  IEEE International Conference on Computer Vision (ICCV) (2017),
  \url{http://lmb.informatik.uni-freiburg.de/Publications/2017/TDB17b}

\bibitem{topnet2019}
Tchapmi, L.P., Kosaraju, V., Rezatofighi, S.H., Reid, I., Savarese, S.: Topnet:
  Structural point cloud decoder. In: The IEEE Conference on Computer Vision
  and Pattern Recognition (CVPR) (2019)

\bibitem{wang2015voting}
Wang, D.Z., Posner, I.: Voting for voting in online point cloud object
  detection. In: Robotics: Science and Systems. vol.~1, pp. 10--15607 (2015)

\bibitem{wang2018sgpn}
Wang, W., Yu, R., Huang, Q., Neumann, U.: Sgpn: Similarity group proposal
  network for 3d point cloud instance segmentation. In: Proceedings of the IEEE
  Conference on Computer Vision and Pattern Recognition. pp. 2569--2578 (2018)

\bibitem{wang2019associatively}
Wang, X., Liu, S., Shen, X., Shen, C., Jia, J.: Associatively segmenting
  instances and semantics in point clouds. In: Proceedings of the IEEE
  Conference on Computer Vision and Pattern Recognition. pp. 4096--4105 (2019)

\bibitem{xiazamirhe2018gibsonenv}
Xia, F., R.~Zamir, A., He, Z.Y., Sax, A., Malik, J., Savarese, S.: Gibson env:
  real-world perception for embodied agents. In: Computer Vision and Pattern
  Recognition (CVPR), 2018 IEEE Conference on. IEEE (2018)

\bibitem{yang2019learning}
Yang, B., Wang, J., Clark, R., Hu, Q., Wang, S., Markham, A., Trigoni, N.:
  Learning object bounding boxes for 3d instance segmentation on point clouds.
  In: Advances in Neural Information Processing Systems. pp. 6737--6746 (2019)

\bibitem{yi2019gspn}
Yi, L., Zhao, W., Wang, H., Sung, M., Guibas, L.J.: Gspn: Generative shape
  proposal network for 3d instance segmentation in point cloud. In: Proceedings
  of the IEEE Conference on Computer Vision and Pattern Recognition. pp.
  3947--3956 (2019)

\bibitem{yuan2018pcn}
Yuan, W., Khot, T., Held, D., Mertz, C., Hebert, M.: Pcn: Point completion
  network. In: 3D Vision (3DV), 2018 International Conference on (2018)

\bibitem{Zhou_2018_CVPR}
Zhou, Y., Tuzel, O.: Voxelnet: End-to-end learning for point cloud based 3d
  object detection. In: The IEEE Conference on Computer Vision and Pattern
  Recognition (CVPR) (June 2018)

\end{thebibliography}

\newpage
\section*{Supplementary Material}
\setcounter{section}{0}
\renewcommand\thesection{S.\arabic{section}}

\section{Controlled experiments and analysis}

In this section, we perform a detailed analysis of GSDN through various controlled experiments. For all experiments, we use the same network architecture, and train and validate the model on the ScanNet dataset~\cite{dai2017scannet}. We use the same hyperparameters for all experiments except for one control variable and train all networks for 60k iterations. Note that the performance of the networks trained for 60k iterations is lower than that of networks trained for 120k iterations reported on the main paper.

\subsection{Balanced cross entropy loss}

One of the main challenges we face during training GSDN is the heavy class imbalance of the sparsity and anchor labels. Such class imbalance is prevalent in object detection and we adopt the balanced cross entropy, one of the well-studied techniques that mitigate various problems associated with class imbalance, for sparsity and anchor prediction. In this section, we demonstrate the effectiveness of the balanced cross entropy loss by comparing it with the network trained with the regular cross entropy loss for sparsity and anchor prediction. We present the object detection result on the ScanNet validation set in Table~\ref{tab:balanced}. The balanced cross entropy loss improves the performance of our network, especially the sparsity prediction. This is due to the nature of our generative sparse tensor decoder which adds cubically growing coordinates from all surface voxels, most of which need to be pruned except for a few points that contain target anchor boxes.

\begin{table}[]
    \centering
    \begin{tabular}{c|c||cc}
    \toprule
    Sparsity loss & Anchor loss & mAP@0.25 & mAP@0.5 \\ \midrule
    CE  & CE  & 33.1 &  9.26  \\
    BCE & CE  &  50.7 & 25.4  \\ \midrule
    BCE & BCE &   \textbf{57.2}   &   \textbf{29.7}  \\ \bottomrule
    \end{tabular}
    \vspace{1em}
    \caption{The effectiveness of balanced cross entropy (BCE) and cross entropy (CE) losses for generative sparse object detection.}
    \label{tab:balanced}
\end{table}

\subsection{Sparsity pruning confidence}

\begin{figure}
    \centering
    \includegraphics[width=0.32\textwidth]{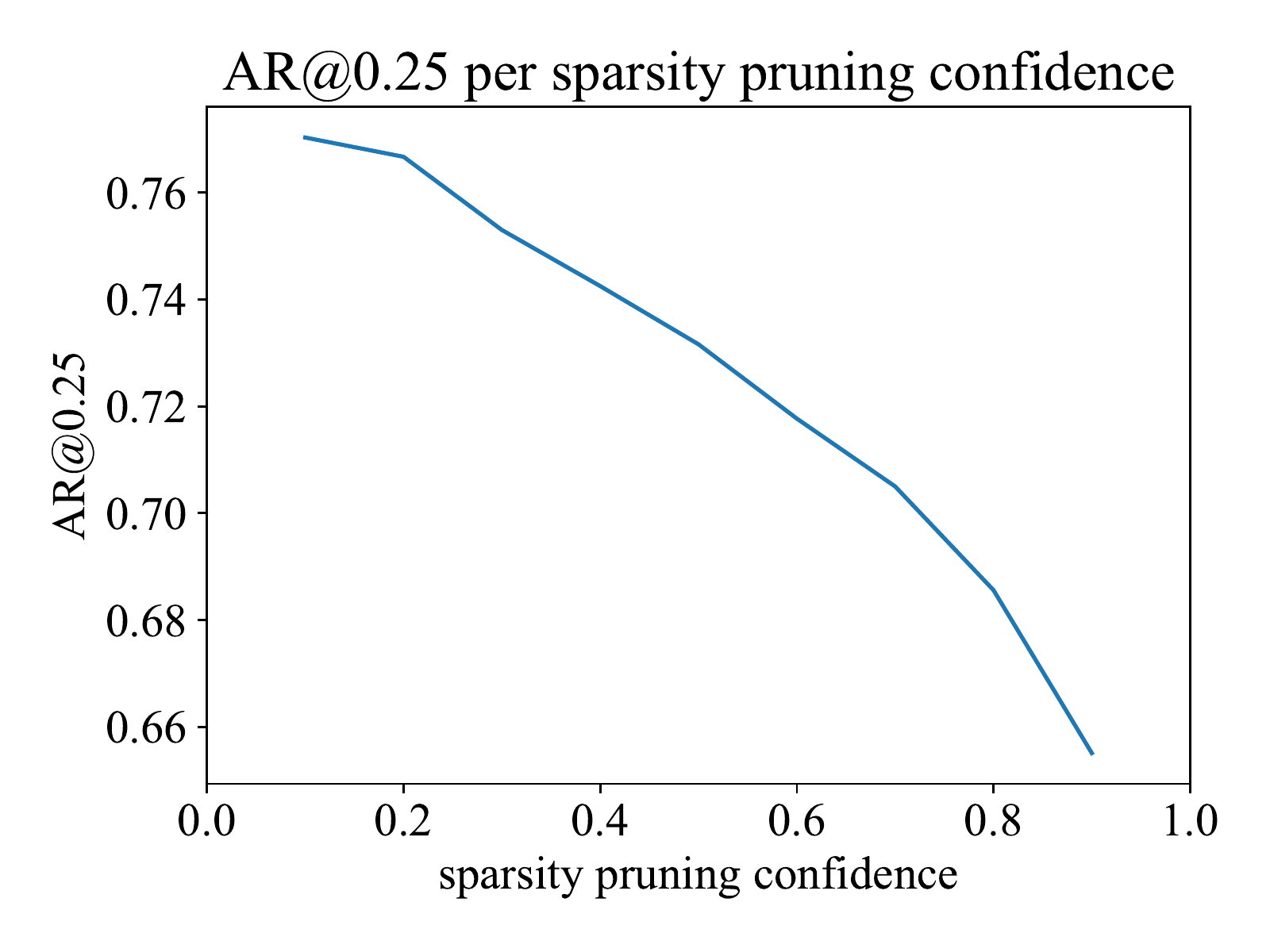}
    \includegraphics[width=0.32\textwidth]{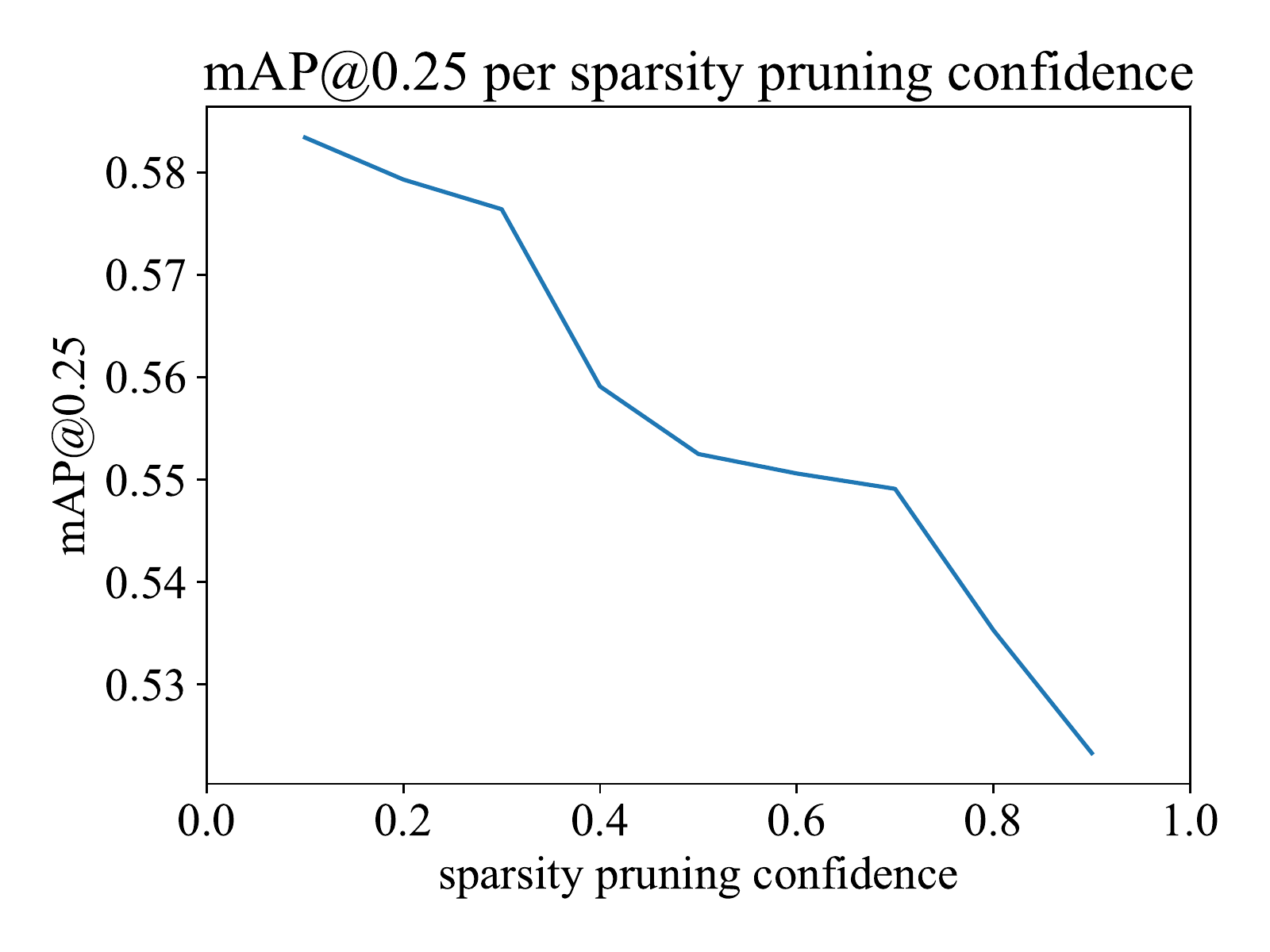}
    \includegraphics[width=0.32\textwidth]{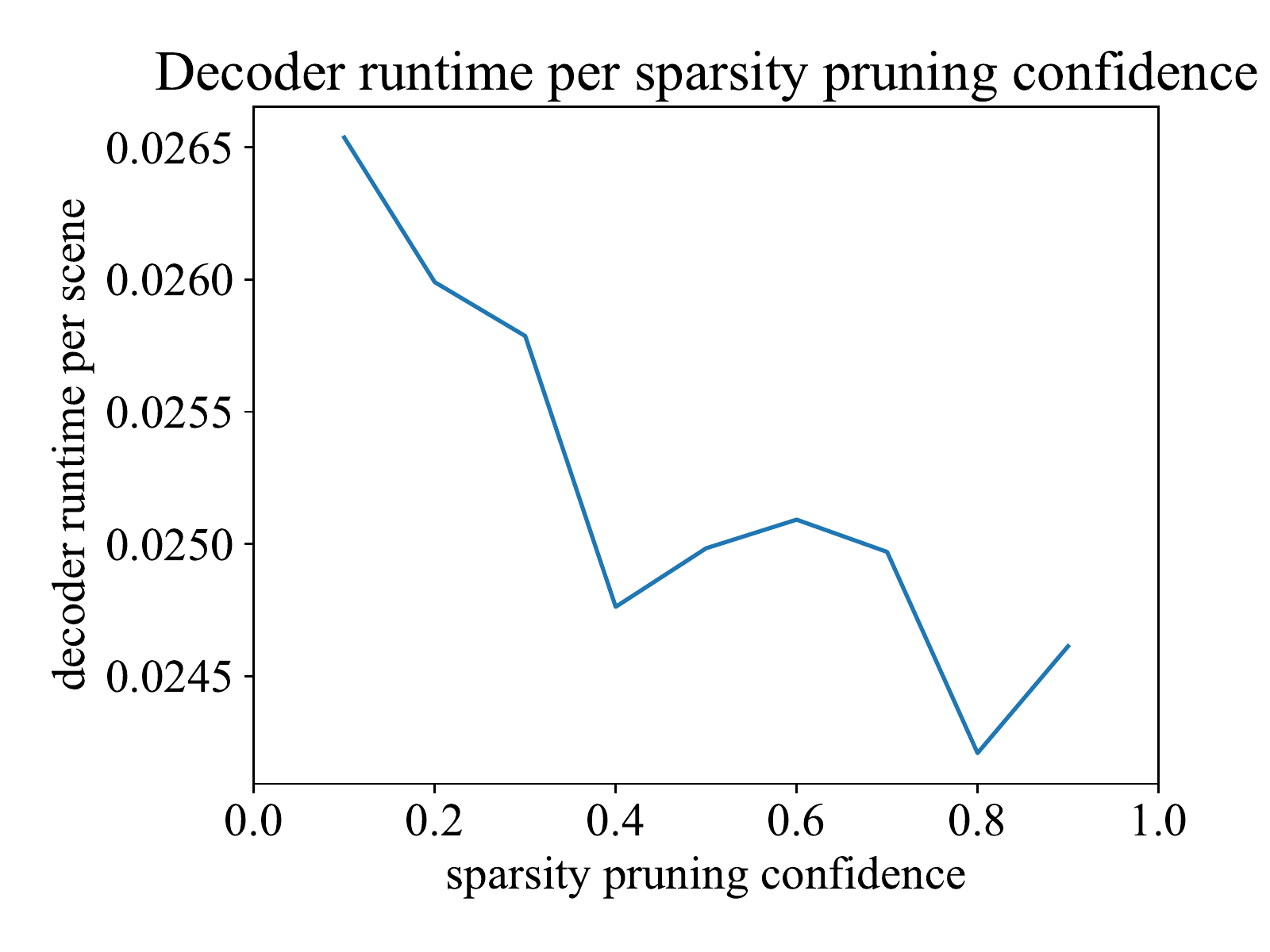}
    \caption{Analysis of the impact of sparsity pruning confidence $\tau$.}
    \label{fig:sparsity_pruning_confidence}
\end{figure}

Our proposed generative sparse detection network predicts more proposals as we lower the sparsity pruning confidence $\tau$ and we found that the threshold $\tau$ has a significant impact on the performance. We analyze the effect of the pruning confidence on average recall, mAP@0.25, and decoder runtime on the ScanNet dataset in Figure~\ref{fig:sparsity_pruning_confidence}. In this experiment, we train three models with $\tau=\{0.1, 0.3, 0.5\}$ and test them on $\tau=\{0.1, 0.2, \ldots, 0.9\}$. For the pruning confidences that do not have corresponding networks, we select the model trained with the closest pruning confidence.

The general trend of Figure~\ref{fig:sparsity_pruning_confidence} is that smaller pruning confidences $\tau$ perform better. Lower pruning confidences lead to more proposals and higher average recall. Also, note that the average precision follows the similar trend, which indicates that the performance is mostly capped by the recall, as shown in the precision/recall curve in the main paper. Lastly, the decoder runs marginally faster as the pruning confidence increases, since it generates fewer proposals.

\subsection{Encoder backbone models}

\begin{figure}
    \centering
    \includegraphics[width=0.5\textwidth]{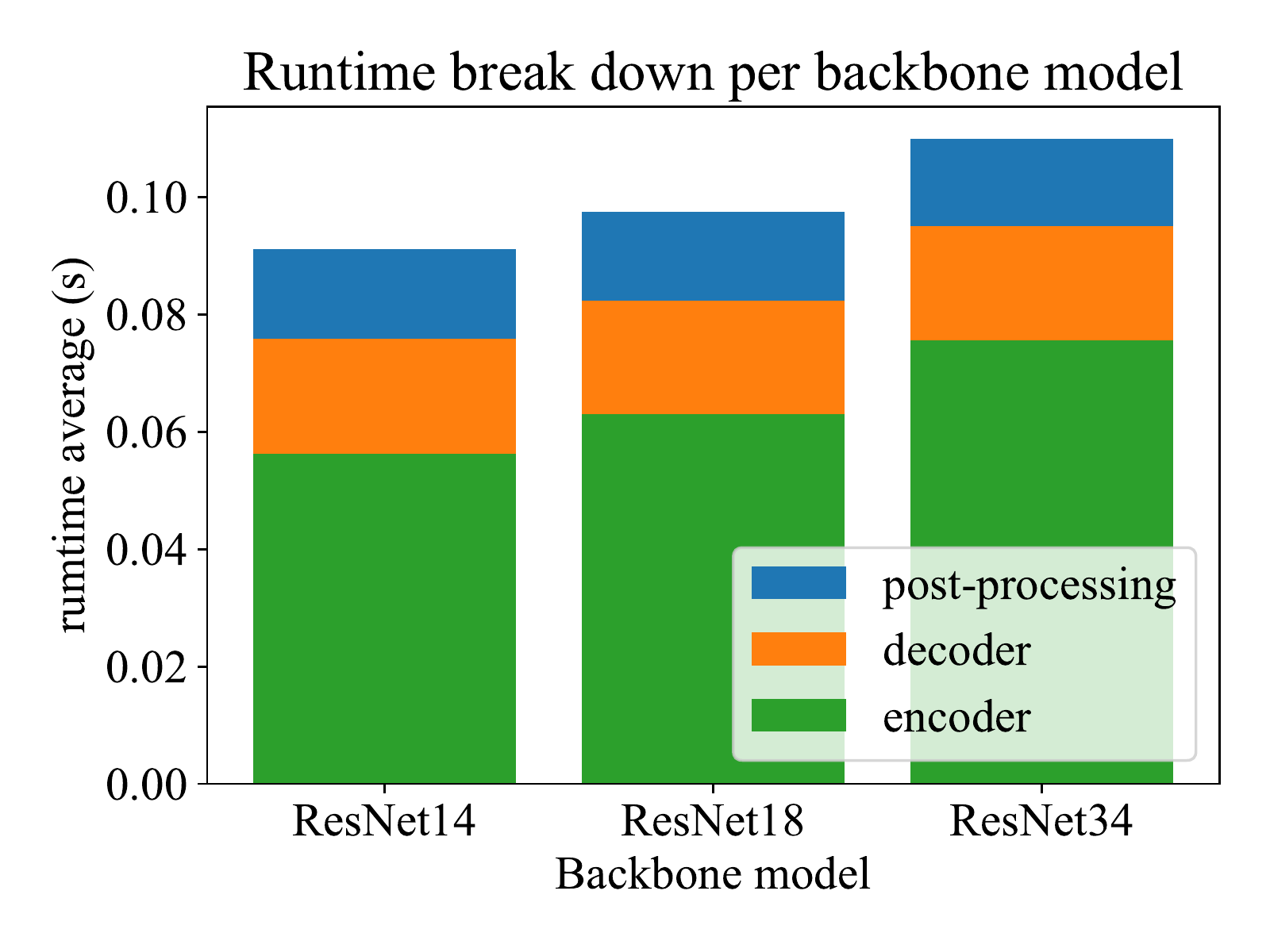}
    \caption{Runtime breakdown of our model with varying backbones.}
    \label{fig:runtime_breakdown}
\end{figure}

We vary the sparse tensor encoder and analyze its impact on performance in Table~\ref{tab:backbone}. As expected, GSDN with deeper encoder performs better. Additionally, we plot the detailed runtime break-down of each component of our proposed model with varying backbones in Figure~\ref{fig:runtime_breakdown}. Overall, the decoder and post-processing time stay almost constant while the encoder dominates the runtime.

\begin{table}[]
    \centering
    \begin{tabular}{c||cc}
    \toprule
    backbone model & mAP@0.25 & mAP@0.5 \\ \midrule
    ResNet14 & 52.1 &  25.4  \\
    ResNet18 & 55.1 &  27.8 \\ \midrule
    ResNet34 & \textbf{57.2}  & \textbf{29.7}   \\ \bottomrule
    \end{tabular}
    \vspace{1em}
    \caption{Analysis of the impact of different backbone models on performance.}
    \label{tab:backbone}
\end{table}

\subsection{Anchor ratios}

We examine the impact of anchor ratios on the performance of our model in Table~\ref{tab:anchor_ratios}. Overall, mAP@0.25 (mAP with IoU threshold of 0.25) improves marginally as we use more anchors. However, the improvement of mAP@0.5 with more anchors is significant. mAP@0.5 considers a prediction box with intersection-over-union greater than 0.5 with the corresponding ground-truth box to be positives. In other words, it requires approximately 80\% overlap between a prediction and the ground-truth box for each of the three axes for the prediction to be positive. Thus, more anchors allow the network to capture various ground truth boxes more accurately and mAP@0.5 improves significantly.

\begin{table}[]
    \centering
    \begin{tabular}{c||cc}
    \toprule
    anchor ratios & mAP@0.25 & mAP@0.5  \\ \midrule
    $a_r \in \{1\}$ & 56.3 &  22.7   \\
    $a_r \in \{1, 4, \frac{1}{4}\}$ & 55.3 &  27.0  \\ \midrule
    $a_r \in \{1, 2, 4, \frac{1}{2}, \frac{1}{4}\}$ & \textbf{57.2}  & \textbf{29.7}  \\ \bottomrule
    \end{tabular}
    \vspace{1em}
    \caption{Analysis of the impact of different anchor ratios on performance.}
    \label{tab:anchor_ratios}
\end{table}

\section{Additional Results}

\subsection{Experiments on the ScanNet dataset}

\begin{figure}[htp!]
\centering
\begin{tabular}{cccc}
G.T. & Hou~\etal~\cite{hou20193d} & Qi~\etal~\cite{qi2019deep} & Ours \\
\includegraphics[width=.21\textwidth]{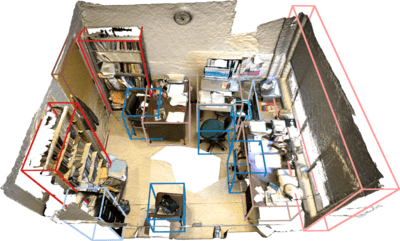} &
\includegraphics[width=.21\textwidth]{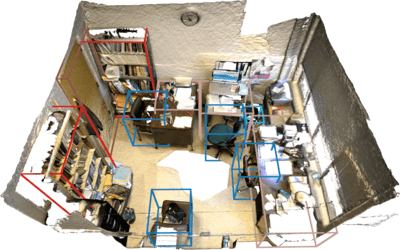} &
\includegraphics[width=.21\textwidth]{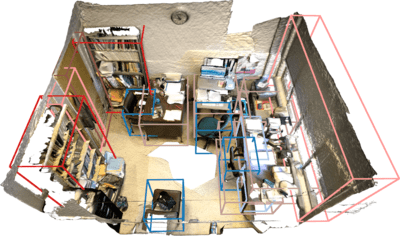} &
\includegraphics[width=.21\textwidth]{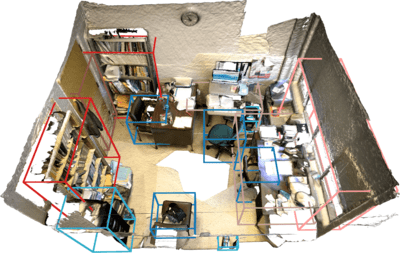} \\
\includegraphics[width=.24\textwidth]{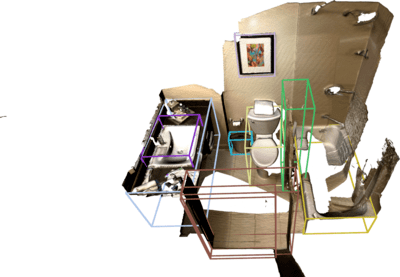} &
\includegraphics[width=.24\textwidth]{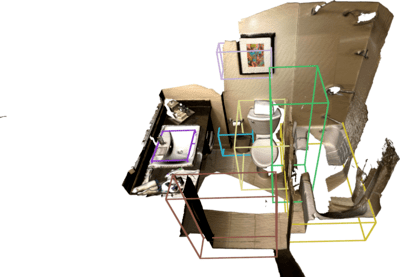} &
\includegraphics[width=.24\textwidth]{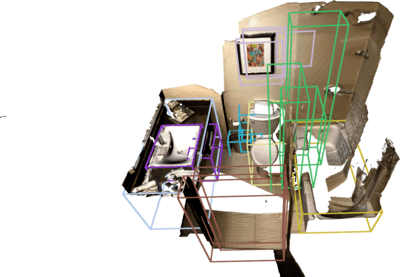} &
\includegraphics[width=.24\textwidth]{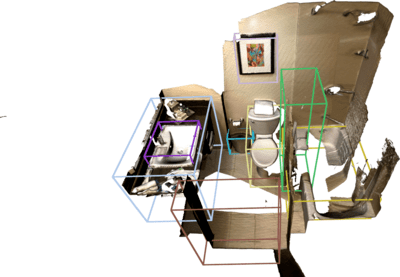} \\
\includegraphics[width=.24\textwidth]{assets/scannet/viz_scene0426_03_gt.png} &
\includegraphics[width=.24\textwidth]{assets/scannet/viz_scene0426_03_3dsis.png} &
\includegraphics[width=.24\textwidth]{assets/scannet/viz_scene0426_03_votenet.png} &
\includegraphics[width=.24\textwidth]{assets/scannet/viz_scene0426_03_ours.png} \\
\includegraphics[width=.24\textwidth]{assets/scannet/viz_scene0565_00_gt.png} &
\includegraphics[width=.24\textwidth]{assets/scannet/viz_scene0565_00_3dsis.png} &
\includegraphics[width=.24\textwidth]{assets/scannet/viz_scene0565_00_votenet.png} &
\includegraphics[width=.24\textwidth]{assets/scannet/viz_scene0565_00_ours.png} \\
\includegraphics[width=.24\textwidth]{assets/scannet/viz_scene0652_00_gt.png} &
\includegraphics[width=.24\textwidth]{assets/scannet/viz_scene0652_00_3dsis.png} &
\includegraphics[width=.24\textwidth]{assets/scannet/viz_scene0652_00_votenet.png} &
\includegraphics[width=.24\textwidth]{assets/scannet/viz_scene0652_00_ours.png} \\
\includegraphics[width=.24\textwidth]{assets/scannet/viz_scene0693_00_gt.png} &
\includegraphics[width=.24\textwidth]{assets/scannet/viz_scene0693_00_3dsis.png} &
\includegraphics[width=.24\textwidth]{assets/scannet/viz_scene0693_00_votenet.png} &
\includegraphics[width=.24\textwidth]{assets/scannet/viz_scene0693_00_ours.png} \\
\includegraphics[width=.24\textwidth]{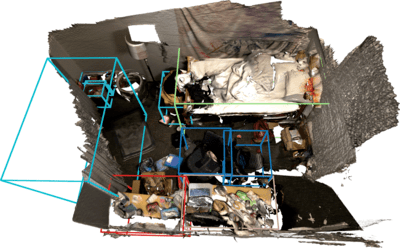} &
\includegraphics[width=.24\textwidth]{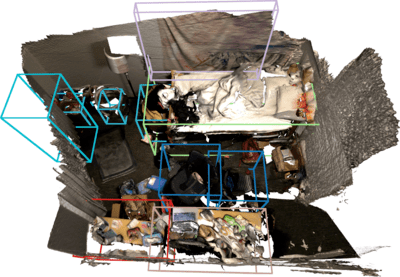} &
\includegraphics[width=.24\textwidth]{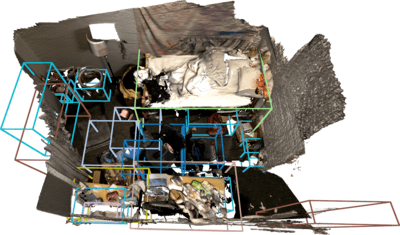} &
\includegraphics[width=.24\textwidth]{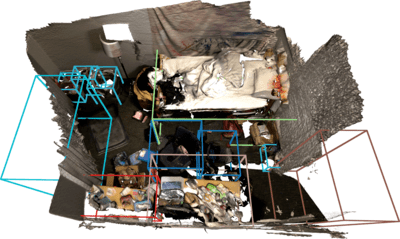} \\
\end{tabular}
\caption{Qualitative object detection results on the ScanNet dataset~\cite{dai2017scannet}.}
\label{fig:scannet_supp_viz1}
\end{figure}

\begin{figure}[htp!]
\centering
\begin{tabular}{cccc}
G.T. & Hou~\etal~\cite{hou20193d} & Qi~\etal~\cite{qi2019deep} & Ours \\
\includegraphics[width=.24\textwidth]{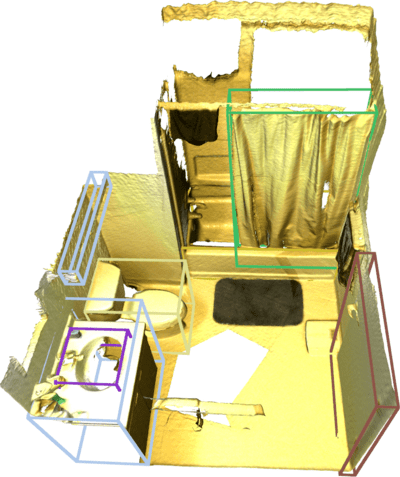} &
\includegraphics[width=.24\textwidth]{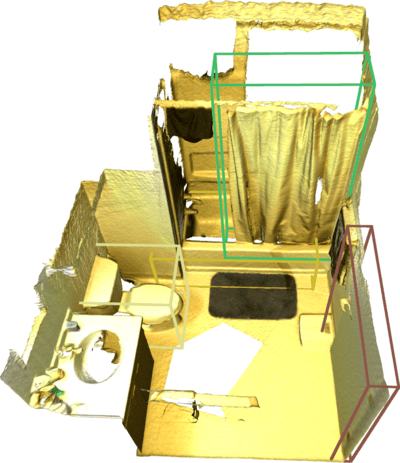} &
\includegraphics[width=.24\textwidth]{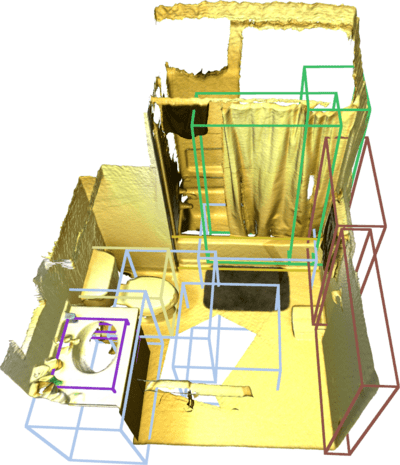} &
\includegraphics[width=.24\textwidth]{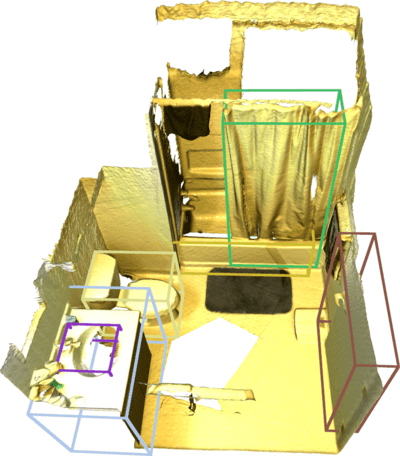} \\
\includegraphics[width=.24\textwidth]{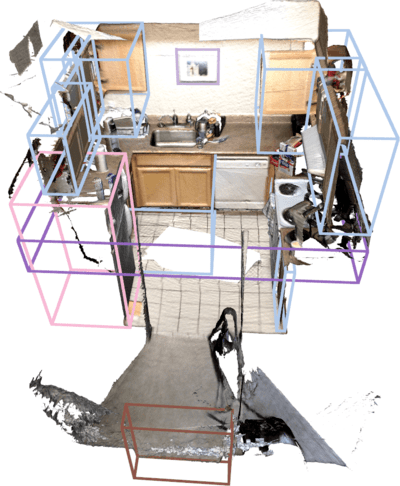} &
\includegraphics[width=.24\textwidth]{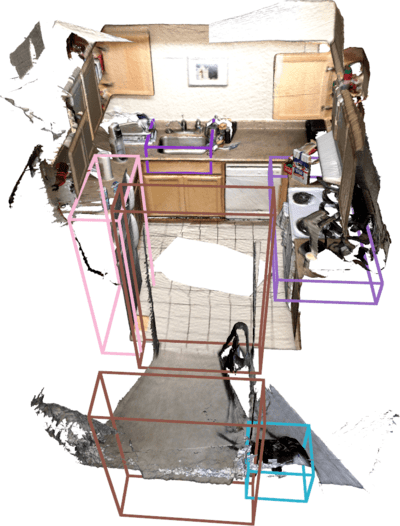} &
\includegraphics[width=.24\textwidth]{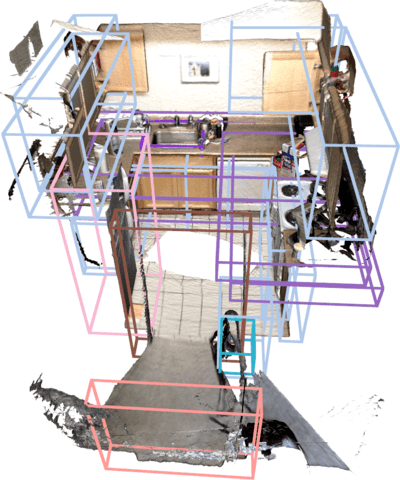} &
\includegraphics[width=.24\textwidth]{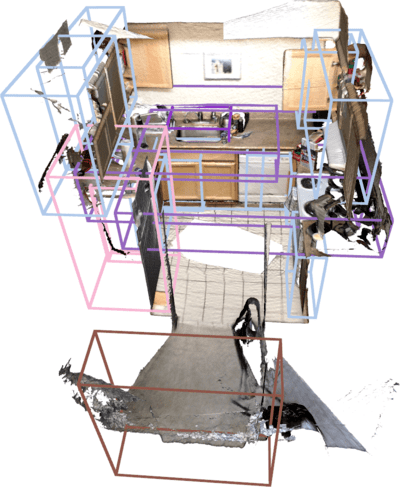} \\
\includegraphics[width=.24\textwidth]{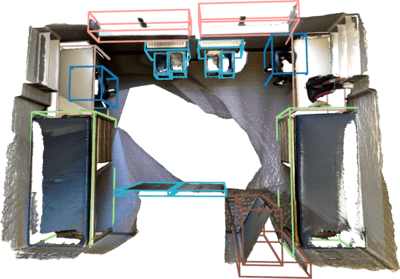} &
\includegraphics[width=.24\textwidth]{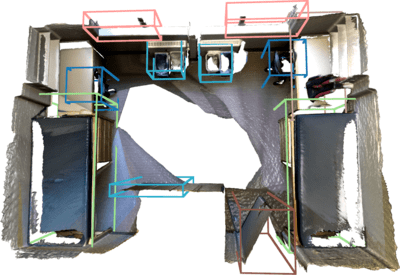} &
\includegraphics[width=.24\textwidth]{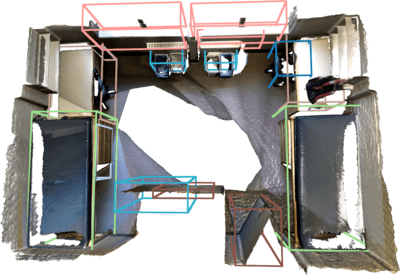} &
\includegraphics[width=.24\textwidth]{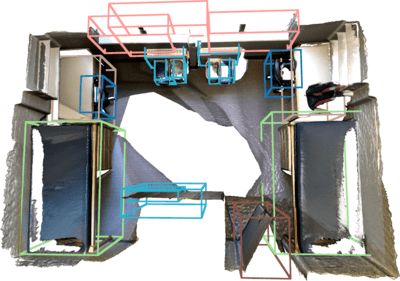} \\
\includegraphics[width=.24\textwidth]{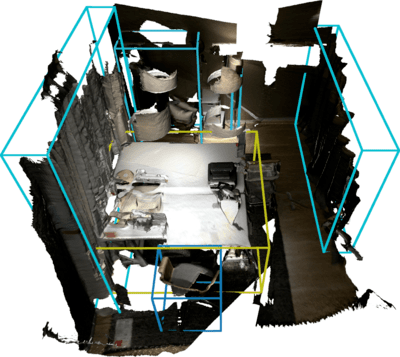} &
\includegraphics[width=.24\textwidth]{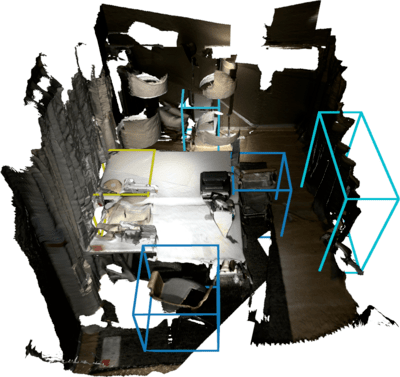} &
\includegraphics[width=.24\textwidth]{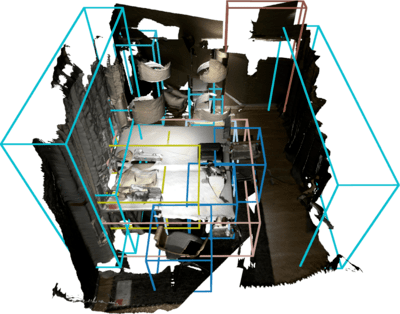} &
\includegraphics[width=.24\textwidth]{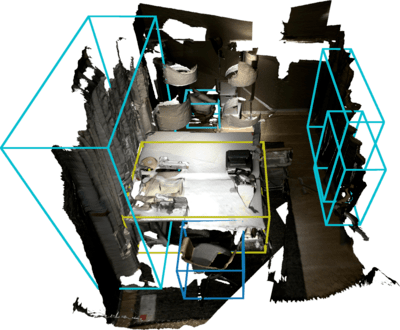} \\
\includegraphics[width=.24\textwidth]{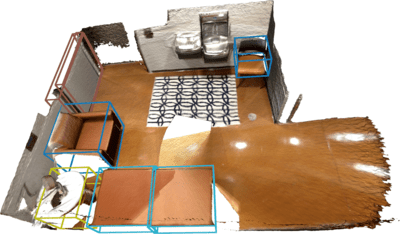} &
\includegraphics[width=.24\textwidth]{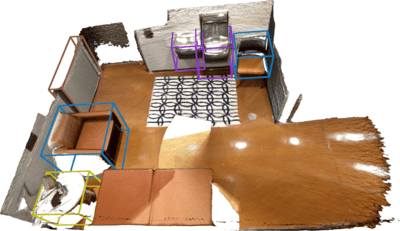} &
\includegraphics[width=.24\textwidth]{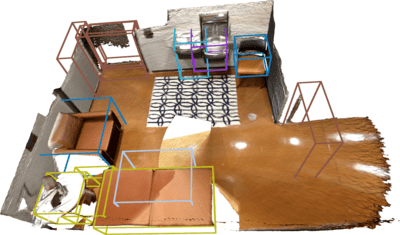} &
\includegraphics[width=.24\textwidth]{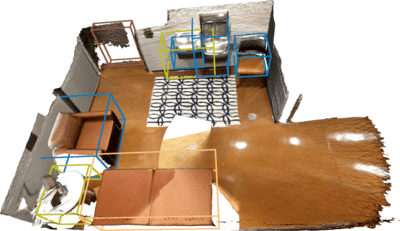} \\
\includegraphics[width=.24\textwidth]{assets/scannet/viz_scene0338_00_gt.png} &
\includegraphics[width=.24\textwidth]{assets/scannet/viz_scene0338_00_3dsis.png} &
\includegraphics[width=.24\textwidth]{assets/scannet/viz_scene0338_00_votenet.png} &
\includegraphics[width=.24\textwidth]{assets/scannet/viz_scene0338_00_ours.png} \\
\end{tabular}
\caption{Qualitative object detection results on the ScanNet dataset~\cite{dai2017scannet}.}
\label{fig:scannet_supp_viz2}
\end{figure}

In Table~\ref{tab:scannet_50}, we report class-wise mAP@0.5 result on the ScanNet v2 validation set. Our method outperforms two-state object detector, Qi~\etal~\cite{qi2019deep}, despite being a single-shot object detector. In Figure~\ref{fig:scannet_supp_viz1} and Figure~\ref{fig:scannet_supp_viz2}, we compare qualitative results on the ScanNet V2 validation set.

\begin{table}[]
\centering
\resizebox{\textwidth}{!}{
\begin{tabular}{l||cccccccccccccccccc|c}
\toprule
              & cab & bed & chair & sofa & tabl & door & wind & bkshf & pic & cntr & desk & curt & fridg & showr & toil & sink & bath & ofurn & mAP \\ \midrule
Hou~\etal~\cite{hou20193d} &  5.06 & 42.19 & 50.11 & 31.75 & 15.12 & 1.38 & 0.00 & 1.44 & 0.00 & 0.00 & 13.66 & 0.00 & 2.63 & 3.00 & 56.75 & 8.68 & 28.52 & 2.55 & 14.60  \\
Hou~\etal~\cite{hou20193d} + 5 views & 5.73 & 50.28 & 52.59 & 55.43 & 21.96 & 10.88 & 0.00 & 13.18 & 0.00 & 0.00 & 23.62 & 2.61 & 24.54 & 0.82 & 71.79 & 8.94 & 56.40 & 6.87 & 22.53 \\
Qi~\etal~\cite{qi2019deep}  & 8.07 & 76.06 & 67.23 & 68.82 & 42.36 & 15.34 & 6.43 & 28.00 & 1.25 & 9.52 & 37.52 & 11.55 & 27.80 & 9.96 & 86.53 & 16.76 & 78.87 & 11.69 & 33.54 \\  \midrule
Ours          & 13.18 & 74.91 & 75.77 & 60.29 & 39.51 & 8.51 & 11.55 & 27.61 & 1.47 & 3.19 & 37.53 & 14.10 & 25.89 & 1.43 & 86.97 & 37.47 & 76.88 & 30.53 & 34.82 \\ \bottomrule
\end{tabular}}
    \vspace{1em}
\caption{Class-wise mAP@0.5 object detection result on the ScanNet v2 validation.}
\label{tab:scannet_50}
\end{table}

\begin{figure}[htp!]
\centering
\begin{tabular}{cccc}
G.T. & Ours & G.T. & Ours \\
\includegraphics[width=.2\textwidth]{assets/stanford/viz_Area5_office_9_gt.png} &
\includegraphics[width=.2\textwidth]{assets/stanford/viz_Area5_office_9_ours.png} &
\includegraphics[width=.2\textwidth]{assets/stanford/viz_Area5_conferenceRoom_1_gt.png} &
\includegraphics[width=.2\textwidth]{assets/stanford/viz_Area5_conferenceRoom_1_ours.png} \\
\includegraphics[width=.2\textwidth]{assets/stanford/viz_Area5_office_34_gt.png} &
\includegraphics[width=.2\textwidth]{assets/stanford/viz_Area5_office_34_ours.png} &
\includegraphics[width=.2\textwidth]{assets/stanford/viz_Area5_conferenceRoom_2_gt.png} &
\includegraphics[width=.2\textwidth]{assets/stanford/viz_Area5_conferenceRoom_2_ours.png} \\
\includegraphics[width=.2\textwidth]{assets/stanford/viz_Area5_office_37_gt.png} &
\includegraphics[width=.2\textwidth]{assets/stanford/viz_Area5_office_37_ours.png} &
\includegraphics[width=.2\textwidth]{assets/stanford/viz_Area5_lobby_1_gt.png} &
\includegraphics[width=.2\textwidth]{assets/stanford/viz_Area5_lobby_1_ours.png} \\
\includegraphics[width=.2\textwidth]{assets/stanford/viz_Area5_office_38_gt.png} &
\includegraphics[width=.2\textwidth]{assets/stanford/viz_Area5_office_38_ours.png} &
\includegraphics[width=.2\textwidth]{assets/stanford/viz_Area5_office_10_gt.png} &
\includegraphics[width=.2\textwidth]{assets/stanford/viz_Area5_office_10_ours.png} \\
\includegraphics[width=.2\textwidth]{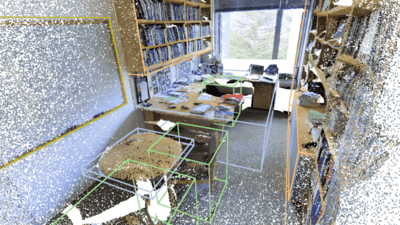} &
\includegraphics[width=.2\textwidth]{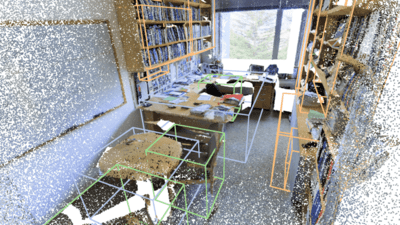} &
\includegraphics[width=.2\textwidth]{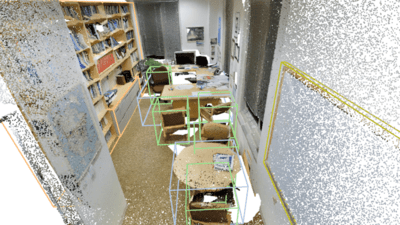} &
\includegraphics[width=.2\textwidth]{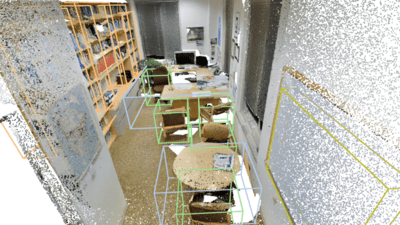} \\
\includegraphics[width=.2\textwidth]{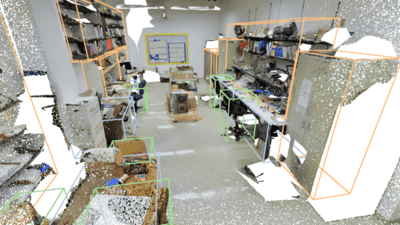} &
\includegraphics[width=.2\textwidth]{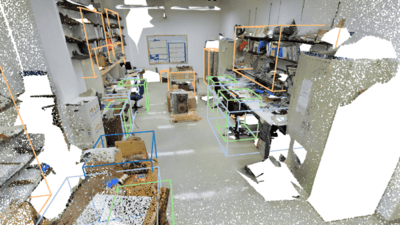} &
\includegraphics[width=.2\textwidth]{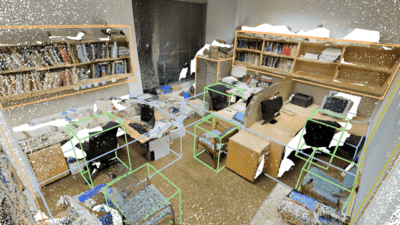} &
\includegraphics[width=.2\textwidth]{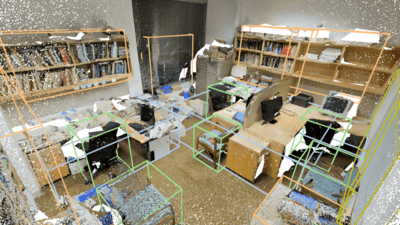} \\
\includegraphics[width=.2\textwidth]{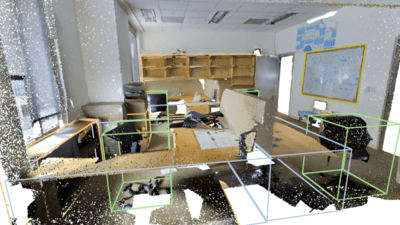} &
\includegraphics[width=.2\textwidth]{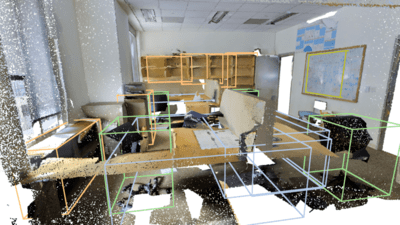} &
\includegraphics[width=.2\textwidth]{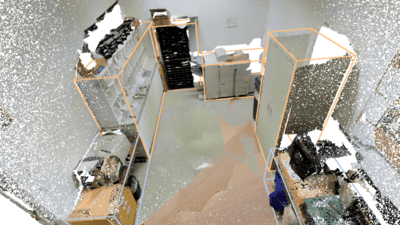} &
\includegraphics[width=.2\textwidth]{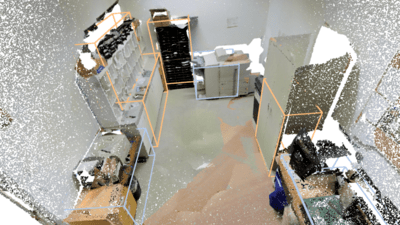} \\
\includegraphics[width=.2\textwidth]{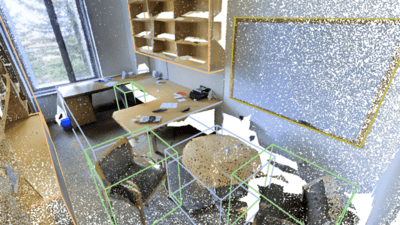} &
\includegraphics[width=.2\textwidth]{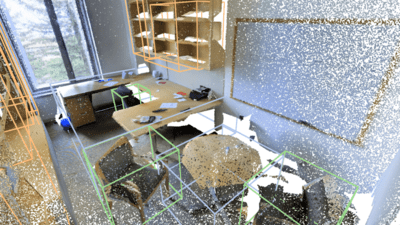} &
\includegraphics[width=.2\textwidth]{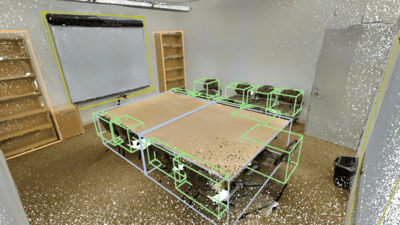} &
\includegraphics[width=.2\textwidth]{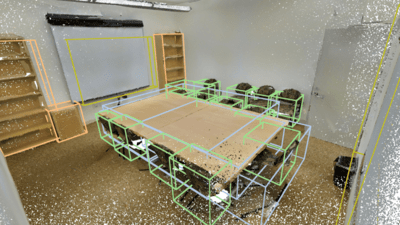} \\
\includegraphics[width=.2\textwidth]{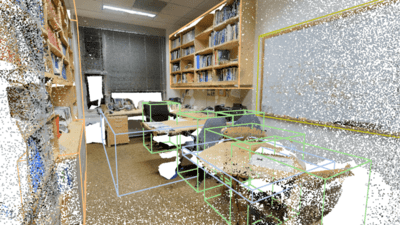} &
\includegraphics[width=.2\textwidth]{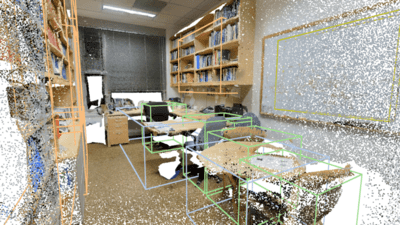} &
\includegraphics[width=.2\textwidth]{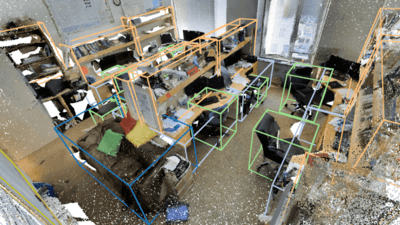} &
\includegraphics[width=.2\textwidth]{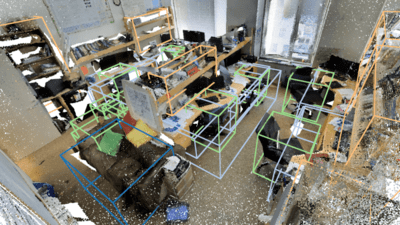} \\
\includegraphics[width=.2\textwidth]{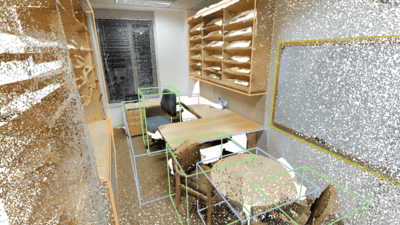} &
\includegraphics[width=.2\textwidth]{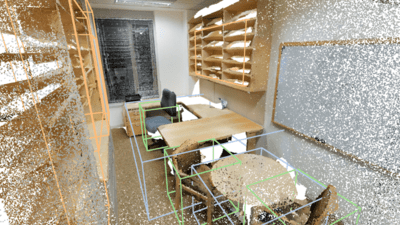} &
\includegraphics[width=.2\textwidth]{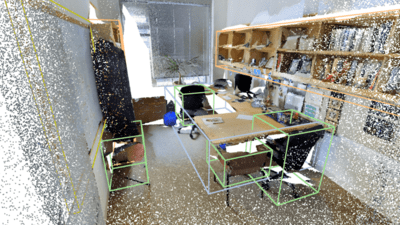} &
\includegraphics[width=.2\textwidth]{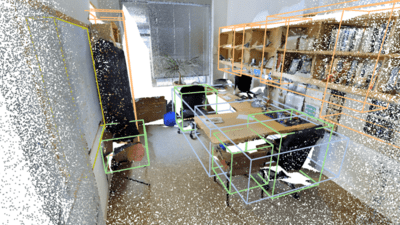} \\
\end{tabular}
\caption{Qualitative object detection results on the S3DIS dataset.}
\label{fig:stanford_supp}
\end{figure}

\subsection{Stanford Large-Scale 3D Indoor Spaces Dataset}

\begin{figure}
    \centering
    \includegraphics[width=0.45\textwidth]{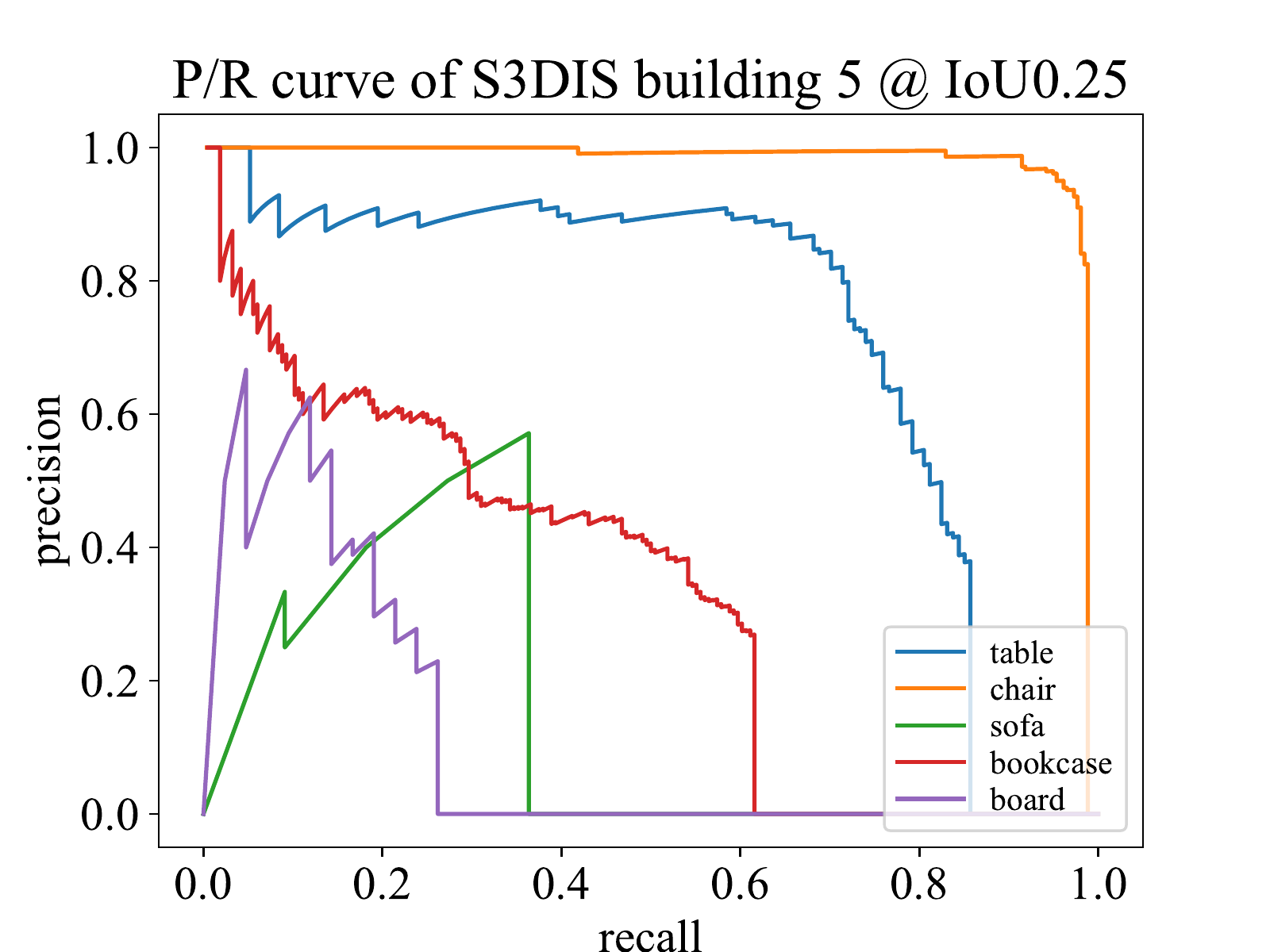} \includegraphics[width=0.45\textwidth]{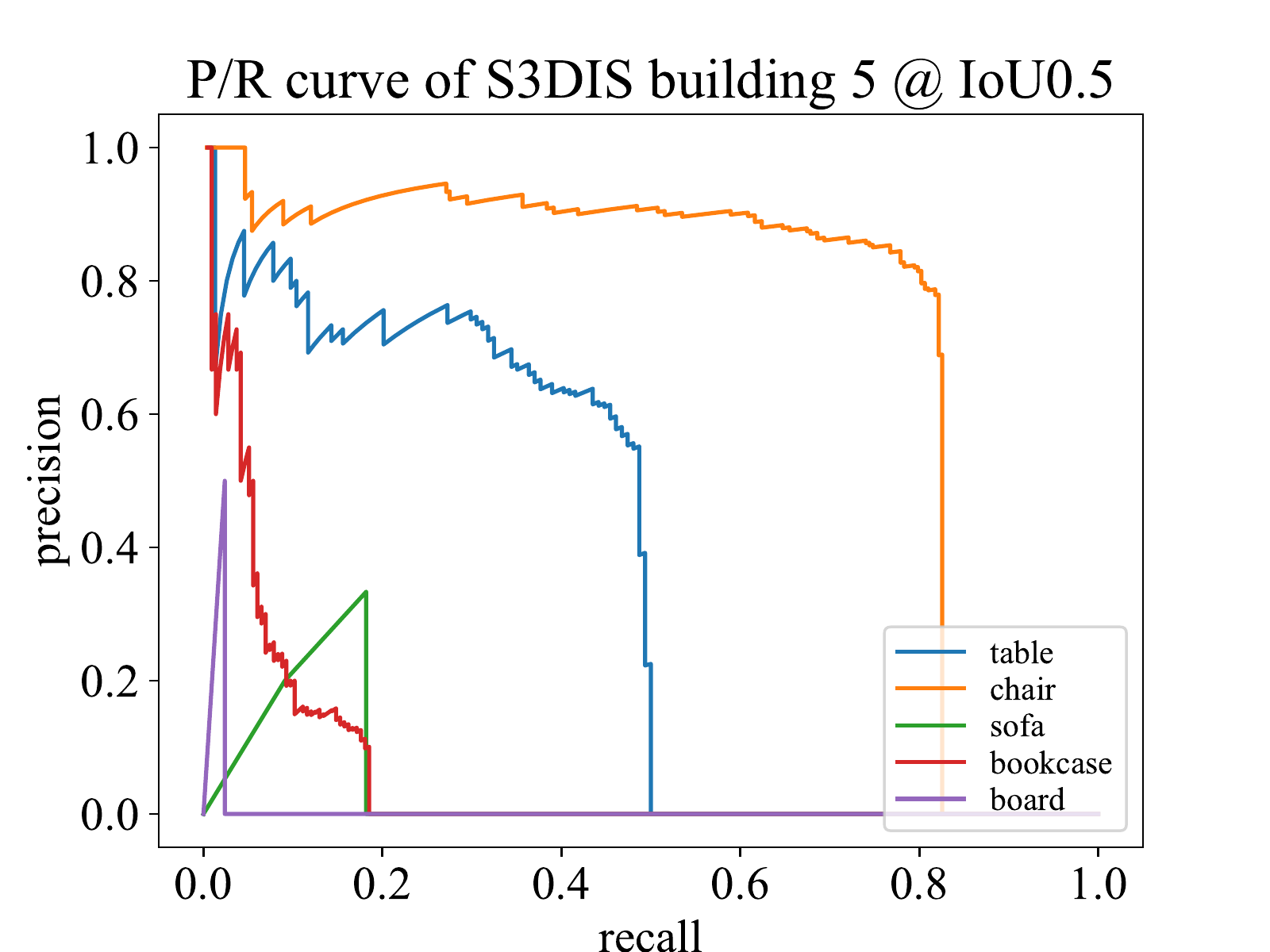}
    \caption{Per-class object detection precision/recall curves of GSDN on the building 5 of the S3DIS dataset.}
    \label{fig:stanford_prcurve}
\end{figure}

We visualize the precision/recall curve of our object detection result on the S3DIS dataset in Figure~\ref{fig:stanford_prcurve}. We observe that certain classes with extreme bounding box ratios such as board and bookcase tend to underperform and have a very low recall.
In Figure~\ref{fig:stanford_supp}, we visualize additional qualitative results of our method on the S3DIS building 5.

\subsection{Gibson environment}

We demonstrate the scalability and generalization capability of our network by testing a model trained on the ScanNet dataset which consists of 3D scans of single-story rooms to the multi-story multi-room building in the Gibson environment~\cite{xiazamirhe2018gibsonenv}. Since our network is fully-convolutional and is translation invariant, our model perfectly generalizes to scenes without extra post-processing such as sliding-window-style cropping and stitching results.

\begin{figure}
    \centering
    \includegraphics[width=0.45\textwidth]{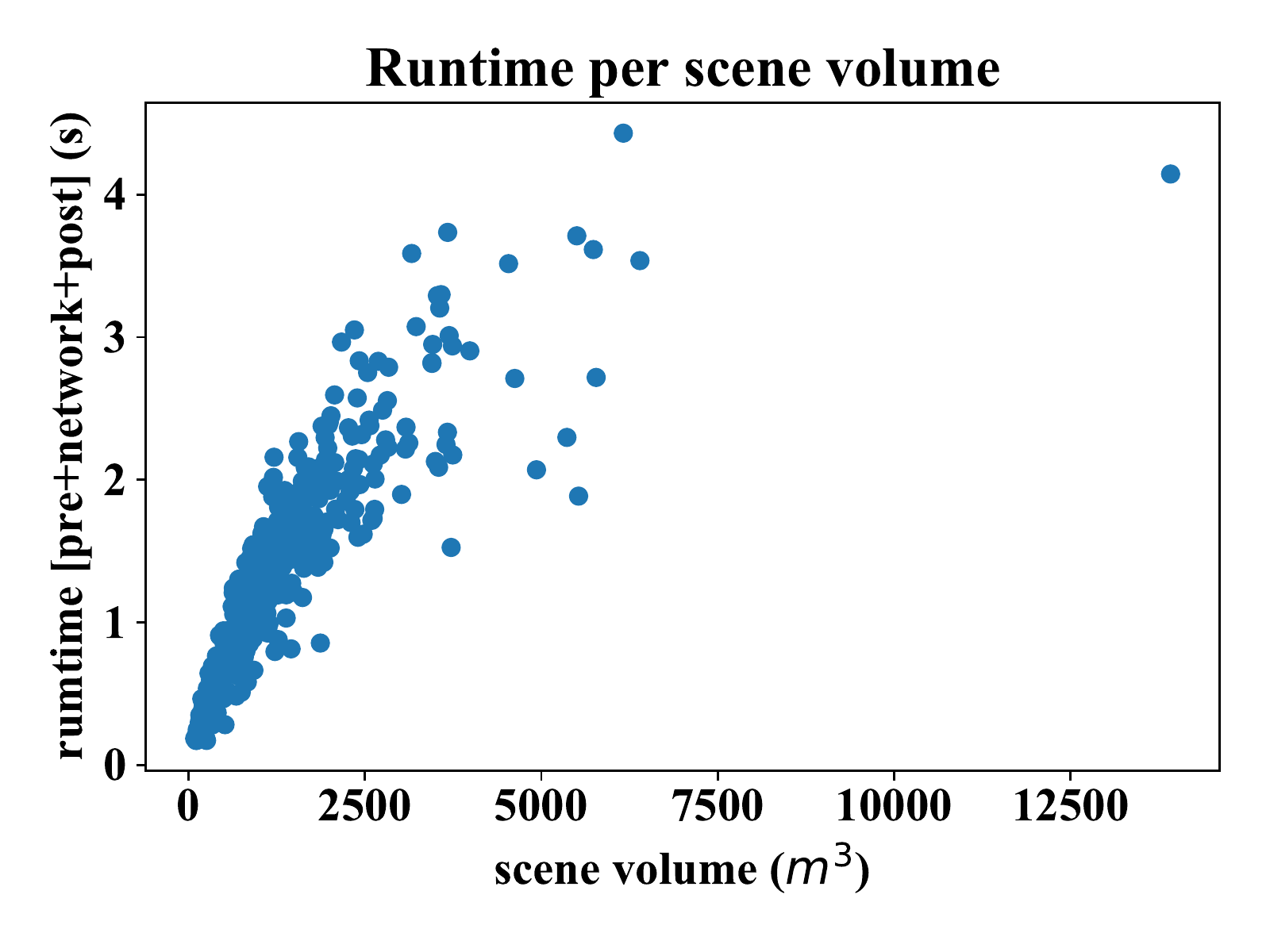} \includegraphics[width=0.45\textwidth]{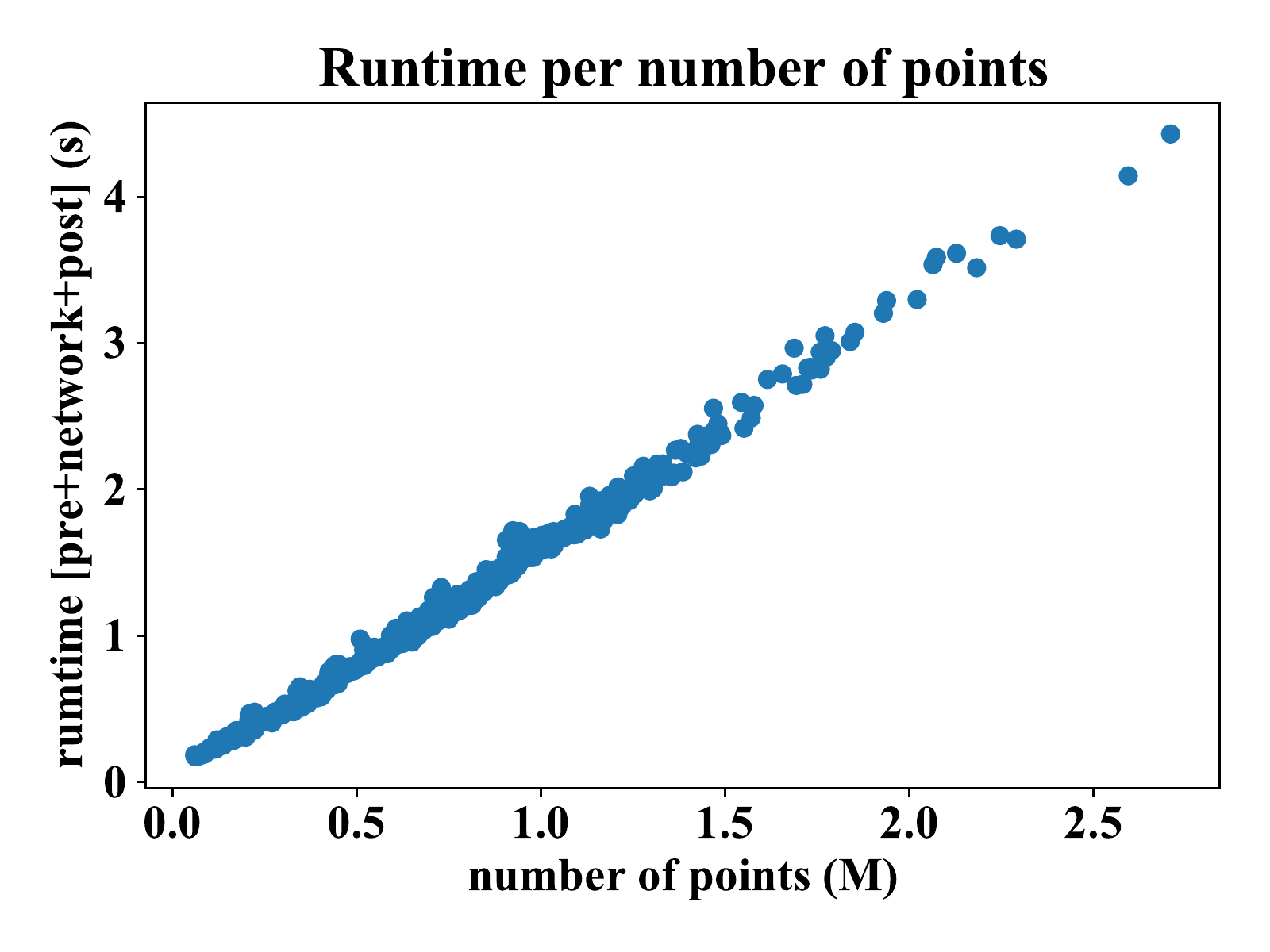} \\
    \includegraphics[width=0.45\textwidth]{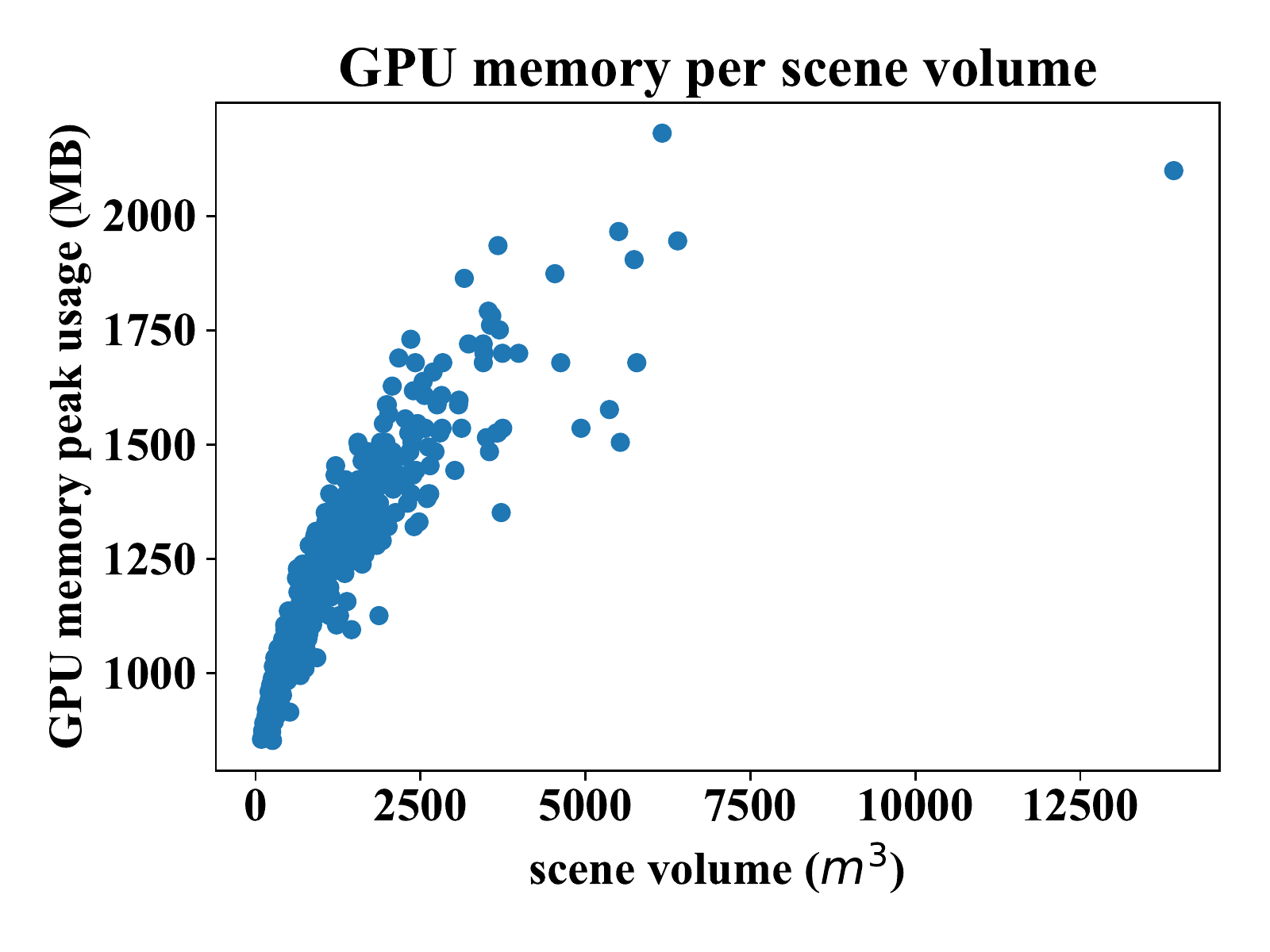}
    \includegraphics[width=0.45\textwidth]{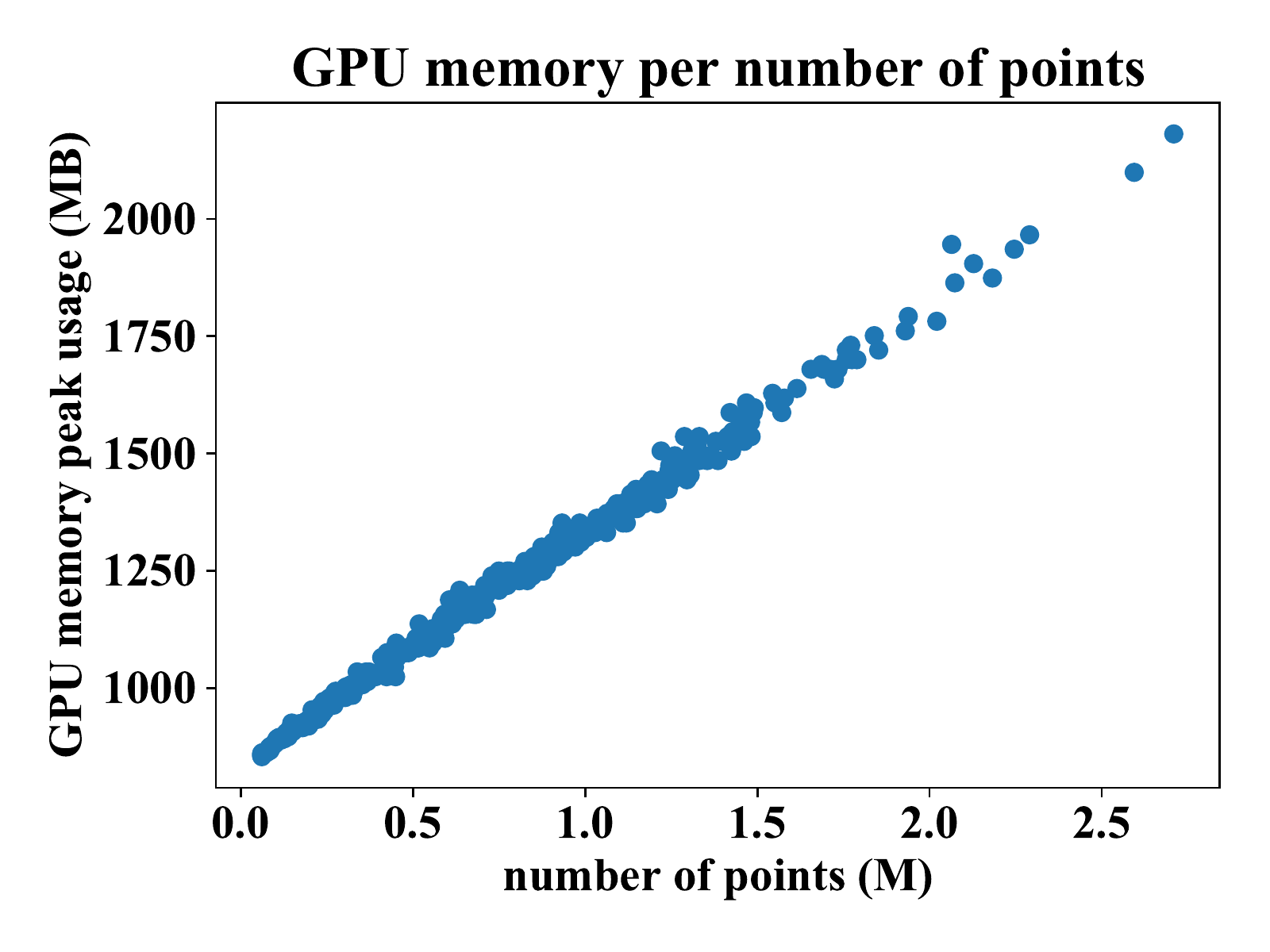}
    \caption{Runtime and peak memory usage analysis on 572 Gibson V2 environments.}
    \label{fig:gibson_analysis}
\end{figure}

\begin{figure}
    \centering
    \includegraphics[width=0.45\textwidth]{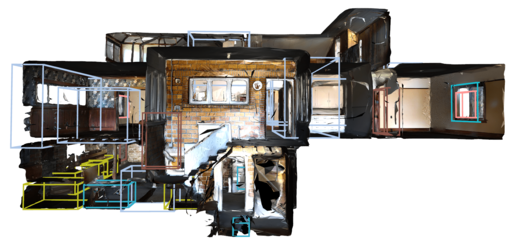} \includegraphics[width=0.45\textwidth]{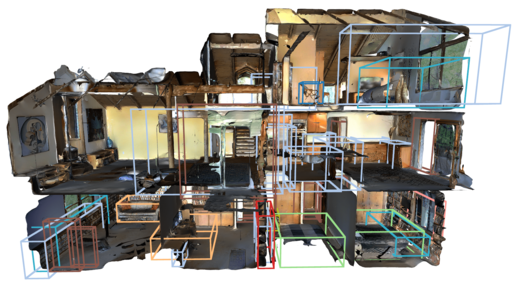} \\
    \includegraphics[width=0.45\textwidth]{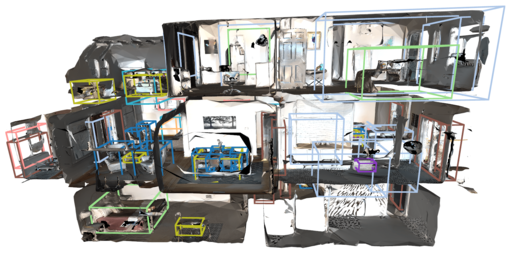}
    \includegraphics[width=0.45\textwidth]{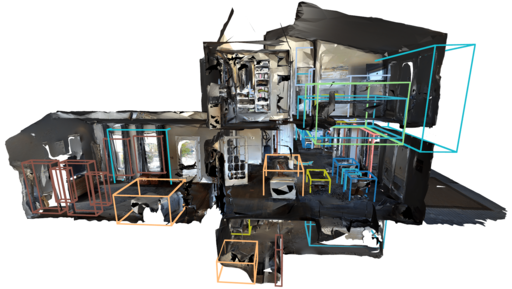} \\
    \includegraphics[width=0.45\textwidth]{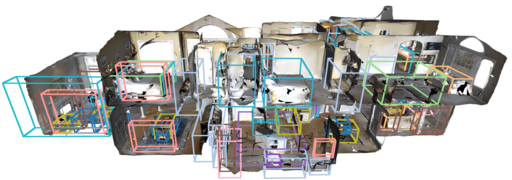}
    \includegraphics[width=0.45\textwidth]{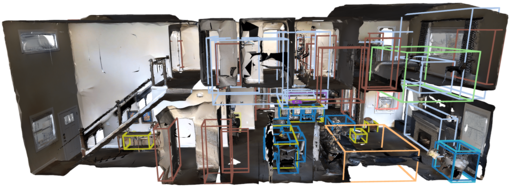}
    \caption{Qualitative object detection results on the Gibson V2 environments.}
    \label{fig:gibson_viz}
\end{figure}

We further analyze the runtime and GPU memory usage of our method on the entire 572 Gibson V2 environments. As shown in Figure~\ref{fig:gibson_analysis}, the runtime and GPU memory usage of our method grows linearly to the number of input points and sublinearly to the volume of the point cloud. This indicates that our method is relatively invariant to the curse of dimensionality. In Figure~\ref{fig:gibson_viz}, we visualize additional qualitative results of our method on the Gibson environment.

\clearpage

%
%
\end{document}